\ifdefined\XeTeXversion\else\pdfoutput=1\fi
\documentclass[10pt, logo, onecolumn, copyright]{nv}

\ifdefined\XeTeXversion\microtypesetup{tracking=false}\fi

\usepackage{graphicx}
\definecolor{nvidiagreen}{HTML}{76B900}

\usepackage{amsmath}
\usepackage{amssymb}
\usepackage{mathtools}
\usepackage{amsthm}
\usepackage{multirow}
\usepackage{algorithm}
\usepackage{algpseudocode}
\usepackage{url}            %
\usepackage{booktabs}       %
\usepackage{caption}        % Separate table/figure captions inside a shared float.
\usepackage{placeins}       % Keep comparison floats ahead of the next experiment.
\usepackage{amsfonts}       %
\usepackage{nicefrac}       %
\usepackage{microtype}      %
\usepackage{xcolor}         %
\usepackage{mathtools}
\usepackage{listings}

\newcommand{\llbodyspacing}{%
  \ifdim\parskip>5pt
    \setlength{\parskip}{5pt plus 1pt minus 1pt}%
  \fi
  \setlength{\textfloatsep}{12pt plus 2pt minus 2pt}%
  \setlength{\floatsep}{8pt plus 2pt minus 2pt}%
  \setlength{\intextsep}{10pt plus 2pt minus 2pt}%
  \setlength{\dbltextfloatsep}{12pt plus 2pt minus 2pt}%
  \setlength{\dblfloatsep}{8pt plus 2pt minus 2pt}%
  \captionsetup{skip=6pt}%
}

\usepackage{xcolor}
\definecolor{pearDark}{RGB}{34,139,34}  % 添加pearDark颜色定义
\definecolor{mygreen}{RGB}{34,139,34}
\definecolor{mylightblue}{RGB}{0,162,230}
\definecolor{deepyellow}{RGB}{255,215,0}
\definecolor{nvgreen}{RGB}{118, 185, 0}

\usepackage{xspace}

\newcommand{\method}{LongLive-Plug\xspace}

\usepackage{mdframed}
\usepackage{color}
\usepackage{xcolor}
\usepackage[utf8]{inputenc} % allow utf-8 input
\usepackage[T1]{fontenc}    % use 8-bit T1 fonts

\usepackage{amsfonts}       % blackboard math symbols
\usepackage{nicefrac}       % compact symbols for 1/2, etc.
\usepackage{microtype}      % microtypography
\usepackage{multirow}
\usepackage{multicol}
\usepackage{tabto}
\usepackage{xspace}
\usepackage{amsmath}
\usepackage{adjustbox}
\usepackage{enumitem}
\usepackage{wrapfig}
\usepackage{times}
\usepackage{verbatim}
\usepackage{amssymb}
\usepackage{mathtools}
\usepackage{caption}
\usepackage{subcaption}
\usepackage{array}
\usepackage{colortbl}
\usepackage{booktabs}
\usepackage{bbm}
\usepackage{makecell}
\usepackage{float}
\usepackage{siunitx}
\usepackage{pifont}
\usepackage{marvosym}
\usepackage{listings}
\usepackage{pdflscape}
\usepackage{footmisc}
\usepackage{url}
\usepackage{tabularx}
\usepackage{arydshln}
\usepackage{hhline}
\usepackage{diagbox}
\usepackage{tcolorbox}
\usepackage[nameinlink]{cleveref}
\usepackage{hyperref}
\usepackage[square,sort,comma,numbers]{natbib} 
\usepackage{fp} % Required for floating-point calculations
\usepackage{authblk}
\usepackage{xspace}
\usepackage{xcolor}         % colors
\usepackage{stfloats}
\definecolor{DeepRed}{RGB}{150,20,20}
\crefname{section}{Sec.}{Sec.}
\crefname{proposition}{Proposition.}{Proposition.}
\crefname{equation}{Eq.}{Eqs.}
\crefname{figure}{Fig.}{Figs.}
\crefname{table}{Tab.}{Tabs.}
\crefname{algorithm}{Algorithm}{Algorithms}
\crefname{appendix}{Appendix}{Appendices}
\Crefname{thm}{Thm}{Thm}
\definecolor{codebg}{RGB}{245, 245, 245}
\definecolor{keywordcolor}{RGB}{0, 0, 153}
\definecolor{commentcolor}{RGB}{34, 139, 34}
\definecolor{stringcolor}{RGB}{163, 21, 21}
\definecolor{numbercolor}{RGB}{128, 128, 128}

\title{LongLive-Plug: Once-for-All Distillation\\for Video Generation}

\runningtitle{LongLive-Plug: Once-for-All Distillation for Video Generation}

\correspondingauthor={}
\newcommand{\llnvauthors}{%
\begin{minipage}[t]{\linewidth}
\centering
{\normalfont\bfseries\fontsize{9}{13}\selectfont
Shuai Yang\textsuperscript{*}\quad
Luozhou Wang\textsuperscript{*}\quad
Wei Huang\quad
ZhiFei Chen\quad
Bohan Zhang\quad
Xiao Fu\\[3pt]
Qianli Ma\quad
Chen-Hsuan Lin\quad
Weian Mao\quad
Bryan Chu\quad
Song Han\quad
Yukang Chen\par}
\vspace{7pt}
{\normalfont\small NVIDIA\par}
\vspace{4pt}
{\normalfont\small
\href{https://github.com/NVlabs/LongLive}{\textcolor{nvidiagreen}{Code}}\qquad
\href{https://nvlabs.github.io/LongLive/LongLive-Plug/}{\textcolor{nvidiagreen}{Project Page}}\qquad
\href{https://huggingface.co/collections/Efficient-Large-Model/longlive-plug}{\textcolor{nvidiagreen}{Models}}\par}
\end{minipage}
}

\begin{abstract}\small
\textbf{Abstract:} Video diffusion models are increasingly developed into specialized models for
diverse downstream tasks, and this development often includes a distillation
stage, for example to accelerate sampling or to improve long-video generation.
This stage is typically repeated for every specialized model.
We introduce \method, a once-for-all distillation framework that learns
reusable capabilities as LoRAs on a base model for training-free, plug-and-play
deployment to compatible downstream models. These capabilities include
single-pass classifier-free guidance,
few-step sampling, and long-context error correction for autoregressive
generation. The adapters remain reusable even when downstream models add
conditioning branches, expand output channels. Despite training at a fixed guidance scale, our dedicated CFG LoRA
provides text guidance control through its inference weight. Combining it with a
few-step LoRA simultaneously preserves few-step generation and CFG
controllability on downstream tasks. We verify training-free deployment on
54 downstream models across three backbone families and eight task categories,
including world modeling, robotics, editing, and multimodal generation.
The approach may support additional compatible models.
Each capability can thus be distilled once per backbone family and reused
without per-target retraining.
\end{abstract}

\newcommand{\llteaserfigure}{%
  \centering
  \includegraphics[width=\linewidth]{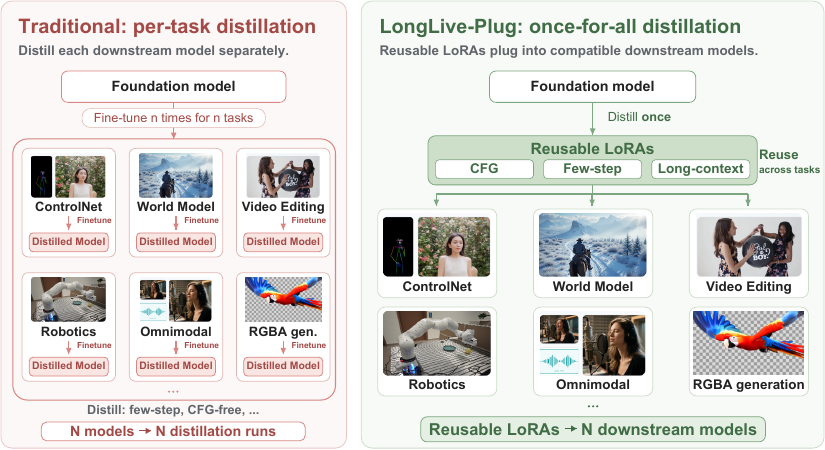}
  \captionof{figure}{\textbf{Once-for-all distillation with plug-and-play deployment.}
  Conventional pipelines separately distill every downstream model (left).
  \method distills CFG, few-step, and long-context capabilities into reusable
  LoRAs on each base model, then transfers them to compatible downstream models
  without target-specific training (right).}
  \label{fig:plug-and-play-distillation}
}

\makeatletter
\let\lloriginalfirststyle\ps@firststyle
\def\ps@firststyle{%
  \lloriginalfirststyle
  \fancyfoot[L]{\copyrightext\hspace{1em}\footerfont $^{*}$ Equal Contributions}%
}
\renewcommand{\maketitle}{%
  \begingroup
  \begin{adjustwidth}{0pt}{0pt}
    \begin{center}
      {\titlefont \@title\par}%
      \vskip11pt
      {\llnvauthors\par}%
      \vskip20pt%
    \end{center}
  \end{adjustwidth}
  \abscontent
  \vskip12pt%
  \llteaserfigure
  \vskip10pt%
  \par\endgroup
  \thispagestyle{firststyle}%
  \markboth{\@runningtitle}{\@runningtitle}%
}
\makeatother

\let\llnvincludegraphics\includegraphics
\newcommand{\llnvfiguresize}[3]{%
  \expandafter\def\csname llnvimage@assets/figures/#1\endcsname{%
    \llnvincludegraphics[width=#2\linewidth,height=#3\textheight,keepaspectratio]{assets/figures/#1}}%
}
\renewcommand{\includegraphics}[2][]{%
  \ifcsname llnvimage@#2\endcsname
    \csname llnvimage@#2\endcsname
  \else
    \llnvincludegraphics[#1]{#2}%
  \fi
}
\llnvfiguresize{plug-and-play-distillation.pdf}{1}{0.46}
\llnvfiguresize{cfg-lora-transferability.pdf}{0.9}{0.27}
\llnvfiguresize{combined-transfer-keyframes.pdf}{0.9}{0.43}
\llnvfiguresize{cfg-guidance-base-keyframes.pdf}{0.9}{0.26}
\llnvfiguresize{cfg-guidance-scope-compact.pdf}{0.82}{0.48}
\llnvfiguresize{ablation-transfer-curves.pdf}{1}{0.22}
\llnvfiguresize{cfg-guidance-base-flower-keyframes.pdf}{0.85}{0.2}
\llnvfiguresize{appendix-cfg-h3-rooftop.pdf}{0.85}{0.35}
\llnvfiguresize{appendix-cfg-h3-villages.pdf}{1}{0.67}

\begin{document}
\maketitle

% Match the title-page overview and fresh-page introduction in the ICLR version.
\clearpage
\begingroup
\llbodyspacing
\section{Introduction}
\label{sec:introduction}

Large-scale video diffusion transformers form reusable foundations for
video generation~\citep{yang2025cogvideox,kong2024hunyuanvideo,wan2025wan},
while downstream applications increasingly depend on specialized models.
They can be adapted to diverse tasks,
including controllable generation and
personalization~\citep{wang2023videocomposer,wei2024dreamvideo}, video
editing~\citep{qi2023fatezero}, and action-conditioned world simulation for
robotics and physical AI~\citep{rigter2024avid,nvidia2025cosmos}.
Developing such a specialized model often includes a distillation stage,
for example to accelerate sampling or to improve long-video generation, and
this stage is typically repeated for every new model, each requiring data
preparation, teacher supervision, and optimization. This repeated cost
motivates a once-for-all
workflow: \textbf{distill reusable capabilities once on a base model and deploy
them across compatible downstream models.}

We introduce \method, a once-for-all distillation framework built on reusable
\emph{functional LoRAs}~\citep{hu2022lora}. Each adapter is learned on a base model and attached
to compatible downstream models while retaining their task-specific weights
(\cref{fig:plug-and-play-distillation}). We consider three capabilities:
\emph{classifier-free guidance (CFG) distillation} combines two-pass guidance
into one model evaluation~\citep{meng2023guided,jensen2025agd} while retaining
continuous control over guidance strength;
\emph{few-step distillation} enables four-step sampling~\citep{yin2024dmd2};
and \emph{long-context distillation} improves long-video quality through
error correction in models that support causal autoregressive (AR) inference.
Once trained on the base model, these adapters support \emph{training-free,
plug-and-play deployment} to compatible downstream models, even when those
models add conditioning branches, or expand output channels.

Making these functional LoRAs transferable raises two further questions.
The first concerns guidance. Existing few-step distillation methods, such as
CausVid and Self Forcing~\citep{yin2025causvid,huang2025selfforcing}, distill
CFG and few-step generation jointly, which fixes guidance at the scale used
during training. Downstream tasks, however, often favor different guidance
strengths, so a single fixed scale cannot serve all of them. We therefore
apply decoupled distillation, which distills CFG and few-step
generation into separate LoRAs. The inference weight of the CFG LoRA then acts
as a guidance dial: scaling it adjusts guidance strength in a near-linear
manner, as we observe empirically and as prior work on scaling fine-tuning
updates suggests~\citep{wortsman2022robust,ilharco2023editing}. Adjusting this
weight while keeping the few-step LoRA fixed tailors guidance to each
downstream task. By contrast, rescaling a jointly distilled LoRA also perturbs
few-step generation and can cause collapse.

The second question is how the training of a functional LoRA affects its
transfer. We report two empirical findings. Adapter rank matters: a small
adapter may fit the base teacher well yet transfer poorly, so source fit alone
does not determine the capacity needed for transfer. The distillation data
also matters: broad T2V prompts on the base model expose the adapter to
diverse generation behavior, and broader prompt coverage improves transfer.
Together, these findings indicate how to learn acceleration that remains
useful beyond the checkpoint on which it was distilled.

The functional-LoRA formulation also extends to \emph{error correction}
for long contexts. We learn this capability through \emph{long-context
distillation}. Using \emph{Streaming Long Tuning}~\citep{yang2026longlive}, we optimize a LoRA
on a causal AR version of the base model using distribution matching
distillation (DMD). The model extends its own generated history, while a
teacher supervises each newly generated short clip. This exposes the adapter
to errors accumulated during extended rollouts and teaches it to sustain
long-video quality. The resulting functional LoRA makes long-context error
correction reusable across compatible downstream models that support causal
AR inference.

We verify training-free deployment on 54 downstream models from three
backbone families: Wan2.1-14B, Wan2.2-TI2V-5B, and MiniMax-H3. They span
eight task categories, including world modeling, robotics, controllable
generation, editing, and multimodal generation. The verified coverage is
listed in \cref{tab:transfer-coverage} in Appendix~\ref{sec:appendix-coverage};
our approach may support additional compatible models.
Quantitative comparisons on SCOPE and Wan2.2-Fun-5B-Control show
improvements over naive four-step sampling and performance competitive with
per-target distillation, without additional downstream training. Further
experiments show that independent CFG control helps transfer to tasks with
different guidance preferences, and that larger adapter ranks and broader
distillation data can improve transfer quality. Separately, transferring
the long-context LoRA to the ReWorld~\citep{chen2026reworld} and
Matrix-Game 3.0~\citep{wang2026matrixgame3} world models improves video quality
during long autoregressive rollouts. These results demonstrate
reusable acceleration and long-context error correction across downstream models.

\section{Related Work}
\label{sec:related-work}

\subsection{Specialized Video Generation}
CogVideoX, Wan, HunyuanVideo, LTX-Video, and Cosmos support diverse video
specializations~\citep{yang2025cogvideox,wan2025wan,kong2024hunyuanvideo,hacohen2025ltxvideo,nvidia2025cosmos}.
Examples include DOVE for super-resolution~\citep{chen2025dove},
VACE for generation and editing~\citep{jiang2025vace}, Matrix-Game 3.0 for
action-conditioned world modeling~\citep{wang2026matrixgame3}, Kiwi-Edit for
guided editing~\citep{lin2026kiwiedit}, HunyuanVideo-Avatar for audio-driven
animation~\citep{chen2025hunyuanvideoavatar}, and Cosmos-Transfer1 for multimodal
control~\citep{nvidia2025cosmostransfer1}.

Acceleration often requires distilling each specialized checkpoint:
FlashMotion and StreamAvatar target trajectory control and avatar
interaction~\citep{li2026flashmotion,sun2025streamavatar}, while LiveEdit and
FlashVSR target streaming editing and super-resolution~\citep{wang2026liveedit,zhuang2026flashvsr}.
D2DF uses one-step consistency distillation for object removal~\citep{chen2026d2df},
DreamDojo uses few-step causal distillation for robot world modeling~\citep{gao2026dreamdojo},
and BiWM applies DMD after camera-control fine-tuning~\citep{rui2026biwm}.
\method instead distills acceleration once per base model for training-free
reuse across compatible descendants.

\subsection{Video Generation Distillation}
Progressive distillation shortens sampling trajectories~\citep{salimans2022progressive},
and distribution matching enables one-step generation~\citep{yin2024dmd}.
VideoLCM uses consistency distillation~\citep{wang2023videolcm}, T2V-Turbo adds
reward feedback~\citep{li2024t2vturbo}, and DOLLAR combines score and consistency
objectives~\citep{ding2025dollar}. LCM-LoRA packages consistency distillation for
reuse across Stable Diffusion fine-tunes~\citep{luo2023lcmlora}.
Guidance distillation merges the two CFG branches into one
pass~\citep{meng2023guided}, while adapter guidance distillation reduces trainable
parameters and examines transfer to image-model derivatives~\citep{jensen2025agd}.
CausVid converts bidirectional teachers into autoregressive generators with
KV caching~\citep{yin2025causvid}; Self Forcing uses autoregressive training
rollouts to reduce the train--test mismatch~\citep{huang2025selfforcing}. We
isolate distilled capabilities in reusable LoRAs for compatible video
specializations: CFG and few-step distillation accelerate inference, while
long-context distillation corrects errors in models that already support AR
inference.
Plug-and-Play Diffusion Distillation transfers a non-LoRA guide network across
fine-tuned image models~\citep{hsiao2024plugplay}.
CASA transfers downstream LoRAs to few-step video models~\citep{wang2026casa},
whereas we transfer distilled capability LoRAs to a broader range of downstream
tasks and model variants.

\section{Method}
\label{sec:method}

\subsection{Preliminaries}
\label{sec:method-setup}

For each backbone family, \method freezes a base model $F_{\theta_0}$ and
distills each capability once into LoRA parameters $\phi$, then reuses them
across compatible downstream models $F_{\theta_\tau}$:
\begin{equation}
    F_{\theta_0}
    \xrightarrow{\text{distill once}} \phi,
    \qquad
    F_{\theta_\tau}
    \xrightarrow[\text{no target training}]{\oplus\phi}
    F_{\theta_\tau\oplus\phi}.
    \label{eq:transfer-objective}
\end{equation}
Here, $\oplus$ adds updates to corresponding layers while retaining the
target's task-specific weights, without downstream training or re-distillation.
We consider three \emph{functional LoRAs}: CFG distillation replaces two-pass
guidance with one conditional evaluation; few-step distillation uses
DMD2~\citep{yin2024dmd2} to enable four-step sampling; and long-context
distillation corrects accumulated errors in models that already support
causal autoregressive (AR) inference.

\subsection{CFG Distillation and Composition}
\label{sec:method-capabilities}
\label{sec:method-design}

\paragraph{CFG distillation.}
For noisy latent $z_t$ at timestep $t$ and condition $c$, let
$v_c=F_{\theta_0}(z_t,t,c)$ and
$v_\varnothing=F_{\theta_0}(z_t,t,\varnothing)$ denote the conditional and
unconditional predictions of the base model. At guidance scale $w$, the
teacher predicts~\citep{ho2022cfg}
\begin{equation}
    v_{\mathrm{cfg}}^{(w)}
    = v_\varnothing+w(v_c-v_\varnothing).
    \label{eq:cfg-teacher}
\end{equation}
At fixed teacher scale $w_{\mathrm{train}}$, we train only the CFG LoRA
$\phi_{\mathrm{cfg}}$ to minimize the expected squared error between its
single-pass conditional output and the teacher's guided prediction, treated
as a fixed target. The backbone, sampling schedule, and attention pattern
remain unchanged.

\paragraph{The CFG LoRA weight as a guidance dial.}
Varying the inference weight $\lambda_{\mathrm{cfg}}$ adjusts the learned
guidance despite fixed-scale training. Under an approximately linear response,
\begin{equation}
\begin{aligned}
    F_{\theta_0\oplus\lambda_{\mathrm{cfg}}\phi_{\mathrm{cfg}}}
    &\approx v_c + \lambda_{\mathrm{cfg}}
    \left(v_{\mathrm{cfg}}^{(w_{\mathrm{train}})}-v_c\right) \\
    &\approx v_{\mathrm{cfg}}^{(\widetilde w)},
    \qquad
    \widetilde w
    =1+\lambda_{\mathrm{cfg}}(w_{\mathrm{train}}-1).
\end{aligned}
\label{eq:cfg-lora-scale}
\end{equation}
Weights $0$ and $1$ recover the conditional model and full distilled adapter,
respectively; larger weights extrapolate. This correspondence is approximate:
nonlinear responses and downstream specialization can change the effective
guidance. We therefore validate control empirically in
\cref{sec:experiments-cfg}.

\paragraph{Decoupled guidance control.}
Scaling a coupled few-step LoRA also changes its learned few-step correction.
To accommodate downstream guidance preferences, we add a separately trained
CFG-only LoRA and adjust $\lambda_{\mathrm{cfg}}$ while fixing the few-step
weight $\lambda_{\mathrm{step}}$ and sampling schedule
(\cref{fig:cfg-lora-transferability}).

At inference, we add the two weighted LoRA updates to each downstream layer:
\begin{equation}
    \widetilde W_\ell^{(\tau)} = W_\ell^{(\tau)}
    + \lambda_{\mathrm{step}}\Delta W_{\ell,\mathrm{step}}
    + \lambda_{\mathrm{cfg}}\Delta W_{\ell,\mathrm{cfg}}.
    \label{eq:cfg-step-composition}
\end{equation}
Here, $W_\ell^{(\tau)}$ is the original target-layer weight;
$\Delta W_{\ell,\mathrm{step}}$ and $\Delta W_{\ell,\mathrm{cfg}}$ are the
separately trained few-step and CFG updates. Merging both updates before
sampling preserves target-specific modules without joint retraining or extra
model evaluations.

\begin{figure*}[t]
    \centering
    \includegraphics[width=\textwidth]{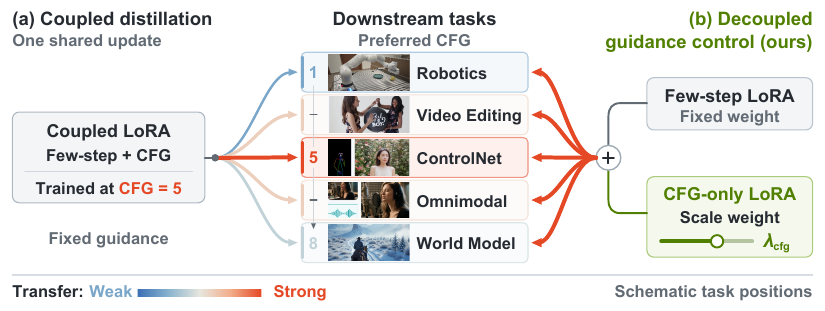}
    \caption{\textbf{Decoupled guidance control through an additional CFG-only LoRA.}
    \textbf{(a)} Scaling a single distilled LoRA changes the entire update,
    including its learned guidance and few-step behavior.
    \textbf{(b)} Adding a separately weighted CFG-only LoRA lets us adjust
    guidance for each target while keeping the few-step adapter weight fixed.
    Arrow colors and widths schematically illustrate guidance compatibility,
    with the warmest coupled-transfer arrow at CFG $5$.
    CFG-only LoRA scaling provides an adjustable guidance control.}
    \label{fig:cfg-lora-transferability}
\end{figure*}

\subsection{Transfer-Oriented Design}
\label{sec:method-transfer-design}

\paragraph{Adapter rank.}
\label{sec:method-rank}
LoRA rank controls the distilled update's capacity. A low-rank adapter may fit
the base teacher yet transfer poorly after downstream specialization. We assess
capacity using both source fit and downstream transfer.
\Cref{sec:experiments-transfer-design} compares ranks under matched training and
deployment protocols, selecting checkpoints by base-model validation.

\paragraph{Distillation data.}
\label{sec:method-data}
We distill CFG and few-step adapters on the base model using broad T2V prompts
covering diverse subjects, scenes, motions, and styles. This exposes the
adapters to varied generation behavior without target training data.
Varying prompt coverage under a fixed teacher isolates prompt diversity;
changing both the teacher and its task data changes the distillation source
jointly. \Cref{sec:experiments-transfer-design} tests how prompt diversity
affects transfer.

\subsection{Long-Context Distillation}
\label{sec:method-long}

To correct errors accumulated during AR rollouts, we train
$\phi_{\mathrm{long}}$ on a frozen causal AR base model using
\emph{Streaming Long Tuning}~\citep{yang2026longlive}. The student generates
each short clip from its cached history, while a pretrained teacher provides
distribution matching distillation (DMD) supervision. Detaching the preceding history keeps gradients local to the
current clip as training rollouts grow longer. The resulting LoRA transfers
to compatible models without target-specific training and improves their
long-context generation quality. It can be applied to models with existing
causal attention.

% Former Section 3.5, retained for reference and excluded from the paper.
% \subsection{Plug-and-Play Deployment}
% \label{sec:method-deployment}
% Once an acceleration LoRA $\phi_k$ has been trained on the base model, we attach
% it to a compatible downstream model by inserting its weighted low-rank update
% into each corresponding layer,
% \begin{equation}
%     \widetilde W_\ell^{(\tau,k)}
%     = W_\ell^{(\tau)}
%     + \lambda_k\Delta W_{\ell,k},
%     \qquad
%     \Delta W_{\ell,k}
%     = \frac{\alpha_k}{r_k}B_{\ell,k}A_{\ell,k},
%     \label{eq:lora-deployment}
% \end{equation}
% where $W_\ell^{(\tau)}$ is the downstream-model weight and $\lambda_k$ is the
% inference weight of adapter $k$. Multiple compatible LoRAs can be combined by
% adding their weighted residuals,
% \begin{equation}
%     \widetilde W_\ell^{(\tau,S)}
%     = W_\ell^{(\tau)}
%     + \sum_{k\in S}\lambda_k\Delta W_{\ell,k}.
%     \label{eq:lora-composition}
% \end{equation}
% In our validated composition,
% $S\subseteq\{\mathrm{cfg},\mathrm{step}\}$: the few-step LoRA enables sampling
% with a short schedule, while $\lambda_{\mathrm{cfg}}$ controls the CFG branch.
% The adapters are applied to the frozen downstream model without target training
% data or optimization. Long-context distillation is evaluated separately on
% compatible models that already support AR inference. Each weighted residual can be merged into
% $W_\ell^{(\tau)}$ before sampling, so the LoRA branches themselves introduce no
% additional inference-time computation.

% NV single-column layout copy; prose and data match sections/04_experiments.tex.
\section{Experiments}
\label{sec:experiments}

\subsection{Comparison of Acceleration Strategies}
\label{sec:experiments-main}

\paragraph{Evaluation tasks.}
We use Wan2.2-TI2V-5B~\citep{wan2025wan} as the foundation model for the main
comparison and evaluate two downstream tasks: world modeling with
SCOPE~\citep{tong2026scope} and ControlNet-based generation with VideoX-Fun's
Wan2.2-Fun-5B-Control~\citep{alibabapai2025wan22fun5bcontrol}.
We evaluate 1{,}378 CrossFPS clips with SCOPE's original input and output
settings and all 600 depth-conditioned PAI-Bench-C cases~\citep{zhou2025paibench}
following its evaluation protocol.
Metrics include FVD~\citep{unterthiner2018fvd},
LPIPS~\citep{zhang2018lpips}, SSIM~\citep{wang2004ssim}, and
DOVER~\citep{wu2023dover}.

\paragraph{Evaluation candidates.}
We compare four candidates: native multi-step inference, naive four-step
sampling, per-target distillation, and \method. Native inference
uses 30 steps for SCOPE and 40 for ControlNet; directly reducing it to four
steps is faster but substantially degrades generation quality. Per-target
DMD2 distillation~\citep{yin2024dmd2} recovers good four-step quality, but
adds substantial per-task data collection, training, and tuning costs.
Our CFG-only and few-step LoRAs are trained on the base model with the broad
T2V data in \cref{sec:method-data}; the few-step adapter uses DMD2 with
CFG-guided teacher supervision.
Combining these adapters enables plug-and-play transfer with
\textbf{zero downstream training cost}, requiring no target data or fine-tuning.
For the quantitative comparisons, \method uses
$(\lambda_{\mathrm{step}},\lambda_{\mathrm{cfg}})=(1,3)$ on
SCOPE (\cref{tab:scope-transfer-results}) and $(1,1)$ on
ControlNet (\cref{tab:control-transfer-results}).

% Quantitative tables float independently of the combined qualitative figure.
\begin{table}[!htbp]
    \centering
    \caption{\textbf{Transfer results on SCOPE.} Metric names, grouping, and
    directions follow Table~1 of SCOPE.
    Native inference uses 30 steps; all accelerated methods use four steps.
    Best values are bolded and second-best values are underlined.}
    \label{tab:scope-transfer-results}
    \fontsize{9}{10}\selectfont
    \renewcommand{\arraystretch}{1.0}
    \setlength{\tabcolsep}{4.4pt}
    \begin{adjustbox}{max width=\textwidth}
    \begin{tabular}{lccccccc}
        \toprule
        & \multicolumn{3}{c}{Visual quality}
        & \multicolumn{2}{c}{Motion quality}
        & \multicolumn{2}{c}{Consistency} \\
        \cmidrule(lr){2-4}\cmidrule(lr){5-6}\cmidrule(lr){7-8}
        Method
        & JEPA $\uparrow$ & FVD $\downarrow$ & LPIPS $\downarrow$
        & Flow $\uparrow$ & Smooth. $\uparrow$
        & Photo. $\downarrow$ & Depth $\downarrow$ \\
        \midrule
        Default 30 step
        & \textbf{0.868} & \textbf{382.9} & \underline{0.651}
        & \textbf{22.11} & \textbf{2.418} & 9.182 & 1.287 \\
        Naive 4 step
        & 0.732 & 805.5 & \textbf{0.628}
        & 11.77 & 2.093 & 8.468 & 1.242 \\
        SCOPE-specific distillation
        & 0.782 & 502.1 & 0.678
        & 16.54 & \underline{2.415} & \underline{5.695} & \textbf{1.203} \\
        \method
        & \underline{0.792} & \underline{478.7} & 0.669
        & \underline{16.87} & 2.399 & \textbf{4.246} & \underline{1.230} \\
        \bottomrule
    \end{tabular}
    \end{adjustbox}
\end{table}

% Quantitative values supplied by the author on 2026-09-16.
\begin{table}[!htbp]
    \centering
    \caption{\textbf{Transfer results on Wan2.2-Fun-5B-Control.}
    All local runs use the same 600 depth-conditioned PAI-Bench-C cases
    and 720P preprocessing. SSIM, F1, si-RMSE, and mIoU measure blurred-RGB,
    edge, depth, and mask similarity; DOVER measures technical quality, and
    LPIPS measures diversity over 3{,}600 videos. All local quantitative runs
    use four steps with CFG disabled. Official scores are from the
    \href{https://huggingface.co/spaces/shi-labs/physical-ai-bench-leaderboard/blob/main/data/conditional_generation-leaderboard.json}{PAI-Bench-C leaderboard}~\citep{zhou2025paibench}, with
    undisclosed settings. Best values across the reported rows are bolded
    and second-best values are underlined.}
    \label{tab:control-transfer-results}
    \fontsize{9}{10}\selectfont
    \renewcommand{\arraystretch}{1.0}
    \setlength{\tabcolsep}{4.4pt}
    \begin{adjustbox}{max width=\textwidth}
    % Fixed decimal precision and right alignment align the decimal points.
    \begin{tabular}{lrrrrrr}
        \toprule
        & \multicolumn{4}{c}{Control fidelity}
        & \multicolumn{2}{c}{Visual quality} \\
        \cmidrule(lr){2-5}\cmidrule(lr){6-7}
        Method
        & \multicolumn{1}{c}{SSIM $\uparrow$}
        & \multicolumn{1}{c}{F1 $\uparrow$}
        & \multicolumn{1}{c}{si-RMSE $\downarrow$}
        & \multicolumn{1}{c}{mIoU $\uparrow$}
        & \multicolumn{1}{c}{DOVER $\uparrow$}
        & \multicolumn{1}{c}{LPIPS $\uparrow$} \\
        \midrule
        Official reported (reference)
        & 0.556 & \textbf{0.106} & 1.819 & \textbf{0.615} & 9.32 & \textbf{0.481} \\
        \midrule
        Naive 4 step
        & \underline{0.560} & 0.094 & 2.135 & 0.582 & 8.90 & 0.264 \\
        ControlNet-specific distillation
        & 0.544 & 0.099 & \textbf{1.515} & 0.595 & \textbf{10.25} & \underline{0.461} \\
        \method
        & \textbf{0.566} & \underline{0.100} & \underline{1.641} & \underline{0.612} & \underline{10.11} & 0.426 \\
        \bottomrule
    \end{tabular}
    \end{adjustbox}
\end{table}

\paragraph{Qualitative and quantitative results.}
\method reduces SCOPE FVD from $805.5$ to $478.7$, comparable to
SCOPE-specific distillation ($502.1$; \cref{tab:scope-transfer-results}).
On ControlNet, it improves all six metrics over naive four-step sampling,
including depth si-RMSE ($2.135$ to $1.641$) and DOVER ($8.90$ to
$10.11$), with metric-dependent trade-offs relative to task-specific
distillation (\cref{tab:control-transfer-results}).
\Cref{fig:main-transfer-comparison} shows clearer scene boundaries and finer
details with preserved control fidelity, demonstrating effective four-step
generation without downstream training.

\subsection{Transfer across Backbones and Tasks}
\label{sec:experiments-coverage}

\paragraph{Coverage beyond the main benchmarks.}
We verify training-free deployment across three backbone families:
Wan2.1-14B, Wan2.2-TI2V-5B~\citep{wan2025wan}, and
MiniMax-H3~\citep{minimax2026h3}. \Cref{tab:transfer-coverage} in
Appendix~\ref{sec:appendix-coverage} lists the 54 verified downstream models:
24 for each Wan backbone and six for H3.
They span eight task categories: world modeling, robotics,
structure-conditioned generation, camera and trajectory control, video
editing and restoration, subject and avatar generation, audio and RGBA
generation, and domain, style, and quality adaptation. Each family reuses
adapters distilled on its own base model without downstream training,
extending capability reuse to full fine-tunes, task LoRAs, and models with
additional conditioning modules. We compare native inference with four-step,
CFG-free inference after attaching the base-distilled LoRA. For models
with 20--50-step native schedules, this reduces denoising steps by
$5$--$12.5\times$. The approach may support additional compatible downstream
models beyond this verified set.

% Author-provided H100 GPU-hours, 2026-09-19: Depth 83.9,
% World model 150.0, Pose 86.8, Robotics sim. 56.1.
% Author-provided base cost, 2026-09-23: 32 GPUs x approximately 2.5 hours
% for 700 iterations = approximately 80 GPU-hours. Source data and generator
% are stored alongside the PDF.
% The ControlNet measurement uses allocation records for job 6624372 through
% its evaluated checkpoint, including model loading and training.
% Share the cost explanation between the two layout formats.
\newcommand{\llcostsummary}{%
\Cref{fig:distillation-cost} tracks cumulative training cost when adding tasks
to Wan2.2-TI2V-5B. Both strategies share a one-time base
distillation cost of approximately $80$ H100 GPU-hours: 700 iterations on
32 GPUs for about 2.5 hours. Task-specific distillation then adds
$83.9$, $150.0$, $86.8$, and $56.1$ H100 GPU-hours for depth-conditioned
generation, world modeling, pose-conditioned generation, and robotics
simulation, respectively: $376.8$ additional GPU-hours and about $456.8$
in total. \method reuses the base-distilled adapters at a fixed cost of about
$80$ GPU-hours, without task-specific data collection. The depth-specific
adapter requires 5{,}000 paired prompts and dynamic depth videos; our transfer
requires no downstream training data.%
}
\newcommand{\llcostcaption}{%
\textbf{Cumulative distillation cost.}
Both curves include the shared base cost of approximately $80$ GPU-hours.%
}
\ifdim\columnwidth<\textwidth
\begin{figure}[tbp]
    \centering
    \includegraphics[width=\linewidth]{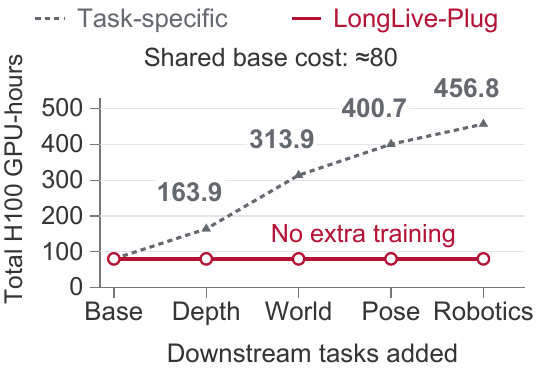}
    \caption{\llcostcaption}
    \label{fig:distillation-cost}
\end{figure}
\paragraph{Cumulative distillation cost.}
\llcostsummary
\else
% Keep the compact explanation and figure together in the available space.
\par\smallskip\noindent
\begin{minipage}[t]{0.57\textwidth}
    \vspace{0pt}
    \textbf{Cumulative distillation cost.}\enspace
    \llcostsummary
\end{minipage}\hfill
\begin{minipage}[t]{0.40\textwidth}
    \vspace{0pt}
    \centering
    \includegraphics[width=\linewidth]{assets/figures/downstream-distillation-cost.pdf}
    \captionof{figure}{\llcostcaption}
    \label{fig:distillation-cost}
\end{minipage}
\par\smallskip
\fi

\begin{figure}[H]
    \centering
    \includegraphics[width=\textwidth]{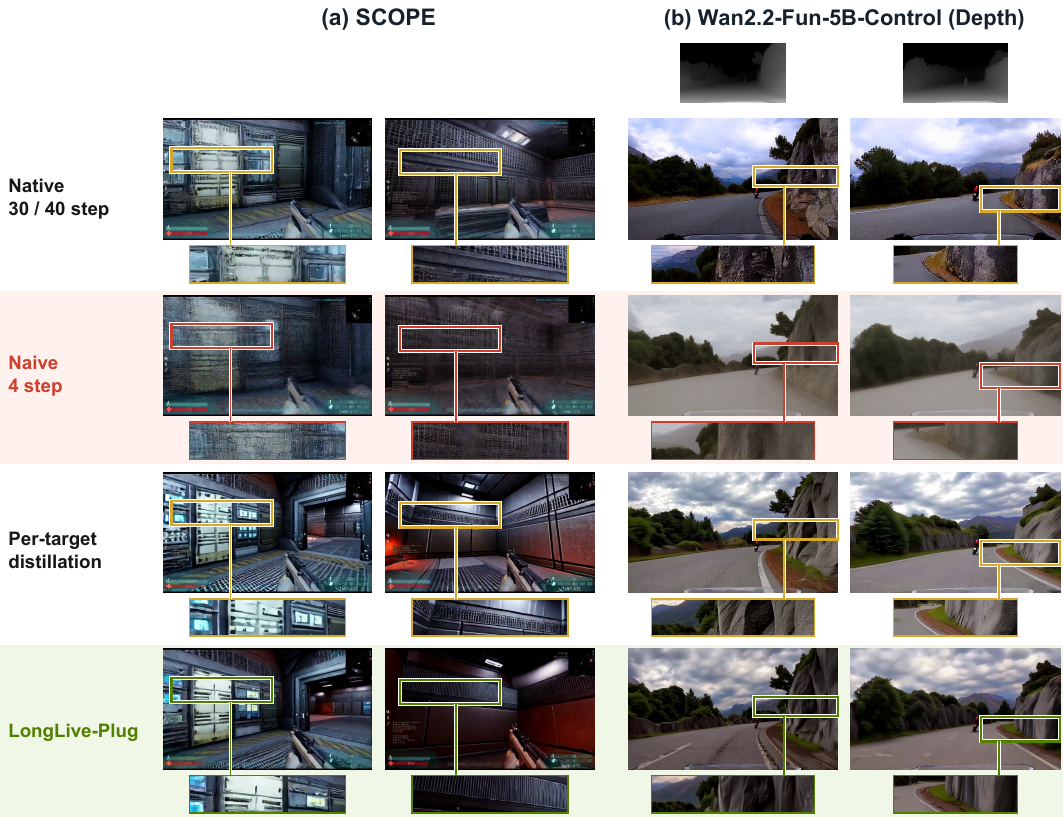}
    \caption{\textbf{Matched transfer comparisons on SCOPE and Wan2.2-Fun-5B-Control (Depth).}
    Two matched frames per task compare native (30/40 steps), naive four-step,
    per-target distilled, and \method outputs.
    Naive four-step sampling produces blurry, low-quality videos, whereas
    \method maintains high visual quality at four steps.
    Depth thumbnails condition ControlNet; boxes and strips show
    matched regions across methods. \method uses an additional base-distilled
    adapter variant. Appendix~\ref{sec:appendix-qualitative} gives full
    four-frame comparisons and alignment details.}
    \label{fig:main-transfer-comparison}
    \label{fig:control-keyframes}
\end{figure}

% Full-width floats need section boundaries in the two-column layout.
% The single-column layout lets text continue in the available space.
\ifdim\columnwidth<\textwidth
\FloatBarrier
\fi
\subsection{CFG Controllability and Composition}
\label{sec:experiments-cfg}

\paragraph{Guidance control with a CFG-only LoRA.}
On Wan2.2-TI2V-5B, we vary only the CFG LoRA weight after distillation at
$w_{\mathrm{train}}=5$, retaining the native 50-step
\href{https://github.com/Wan-Video/Wan2.2/blob/main/wan/utils/fm_solvers_unipc.py}{FlowUniPC}~\citep{zhao2023unipc} schedule.
Raising $\lambda_{\mathrm{cfg}}$ from $1$ to $2$ or $3$ strengthens the milk
splash (\cref{fig:cfg-guidance-base}) at runtime CFG $1$, preserving guidance
control with one conditional evaluation per step.
See Appendix~\ref{sec:appendix-cfg} (\cref{fig:cfg-guidance-base-flower}) for
more cases.

\begin{figure}[!htbp]
    \centering
    \includegraphics[width=\textwidth]{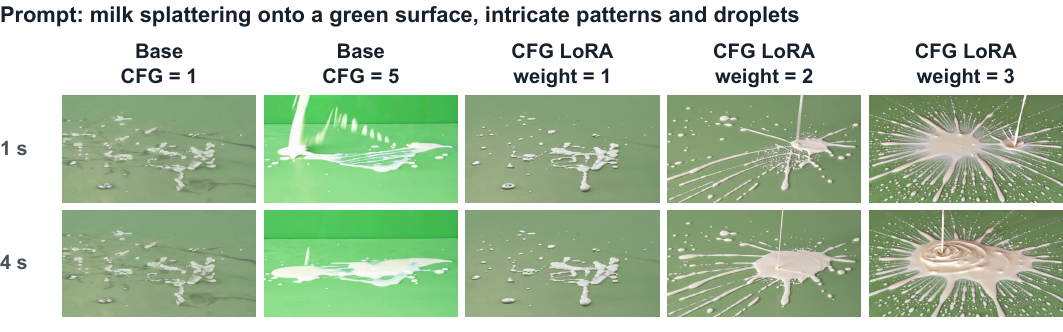}
    \caption{\textbf{Guidance control after CFG-only distillation.}
    The milk-splatter prompt compares native CFG references
    with CFG LoRA weights $1$, $2$, and $3$. All variants use 50 sampling steps;
    every LoRA variant uses the distilled CFG setting and requires only one
    conditional forward pass per step.
    Rows show matched frames at 1 and 4 seconds. Higher LoRA weights follow
    the trend of stronger CFG, producing a more pronounced splash without
    exactly matching native CFG scales.}
    \label{fig:cfg-guidance-base}
\end{figure}

\paragraph{Independent guidance after transfer.}
On SCOPE, a four-step coupled CFG-plus-step LoRA responds
weakly to changes between the Snow Village and Crystal Maze prompts.
Adding a separately weighted CFG LoRA
strengthens the requested snow and crystal attributes as
$\lambda_{\mathrm{cfg}}$ increases from $1$ to $3$ and $5$, while the
few-step weight stays at $1$ (\cref{fig:cfg-guidance-transfer}A).
Matched frames preserve recognizable geometry and the foreground weapon as
text control changes at a fixed few-step weight.

\paragraph{Failure of global LoRA scaling.}
Globally scaling the coupled LoRA fails to provide effective guidance control
on SCOPE. With the checkpoint, prompt, input image, action sequence, seed,
and sampler fixed, increasing the global weight darkens and distorts the
scene, with severe collapse at weight $5$ (\cref{fig:cfg-guidance-transfer}B).
Both SCOPE experiments use four steps with distilled CFG and different
coupled checkpoints. See Appendix~\ref{sec:appendix-cfg} for more cases and
experimental settings.

\begin{figure}[!htbp]
    \centering
    \includegraphics[width=\textwidth]{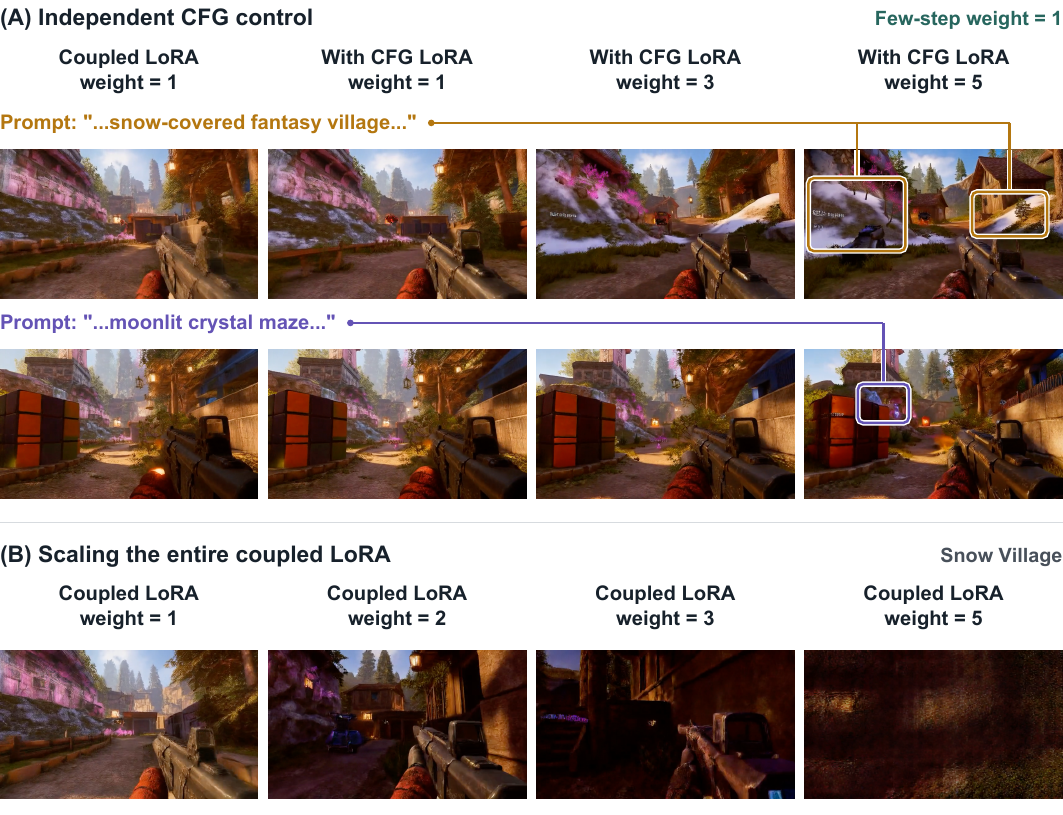}
    \caption{\textbf{Independent CFG control on SCOPE.}
    (A) The coupled LoRA alone at weight $1$ shows little response to the prompts.
    Adding a separately weighted CFG LoRA strengthens the boxed prompt attributes
    while the coupled LoRA weight remains at $1$. Colored lines link each box
    to its corresponding prompt text.
    (B) Scaling the entire coupled LoRA instead degrades generation.
    Frames are matched across weights within each row. All runs use four-step
    sampling with distilled CFG; (A) and (B) use different coupled checkpoints.}
    \label{fig:cfg-guidance-transfer}
\end{figure}

\ifdim\columnwidth<\textwidth
\FloatBarrier
\fi
\subsection{Transfer Design Ablation}
\label{sec:experiments-transfer-design}
\label{sec:experiments-rank}
\label{sec:experiments-data}

\paragraph{Adapter rank.}
Following \cref{sec:method-transfer-design}, we vary rank while fixing the
teacher, prompts, target layers, optimization
budget, and checkpoint-selection rule. Each adapter transfers to SCOPE without
target training and is evaluated by FVD on the full CrossFPS test set.
Across three rank doublings from $16$ to $128$, transfer improves monotonically
by $21\%$ (\cref{fig:rank-ablation}a).

\paragraph{Distillation data.}
We vary prompt diversity with the teacher, rank, target layers, optimization
budget, and number of training lines fixed. Prompt concentration is the mean
pairwise cosine similarity in centred UMT5~\citep{chung2023unimax} embedding space: lower values
indicate broader coverage, while higher values indicate prompts clustered
in one region. The number of distinct prompts co-varies with concentration
within the fixed line budget. Transfer degrades monotonically as diversity
falls, with FVD rising by $12\%$ across the sweep (\cref{fig:data-ablation}b).
At the same training budget, broader prompt coverage thus better supports
transfer to models unseen during distillation.

% Four independent axes, redrawn for readable typography at quarter width.
\begin{figure}[!htbp]
    \centering
    \includegraphics[width=\textwidth]{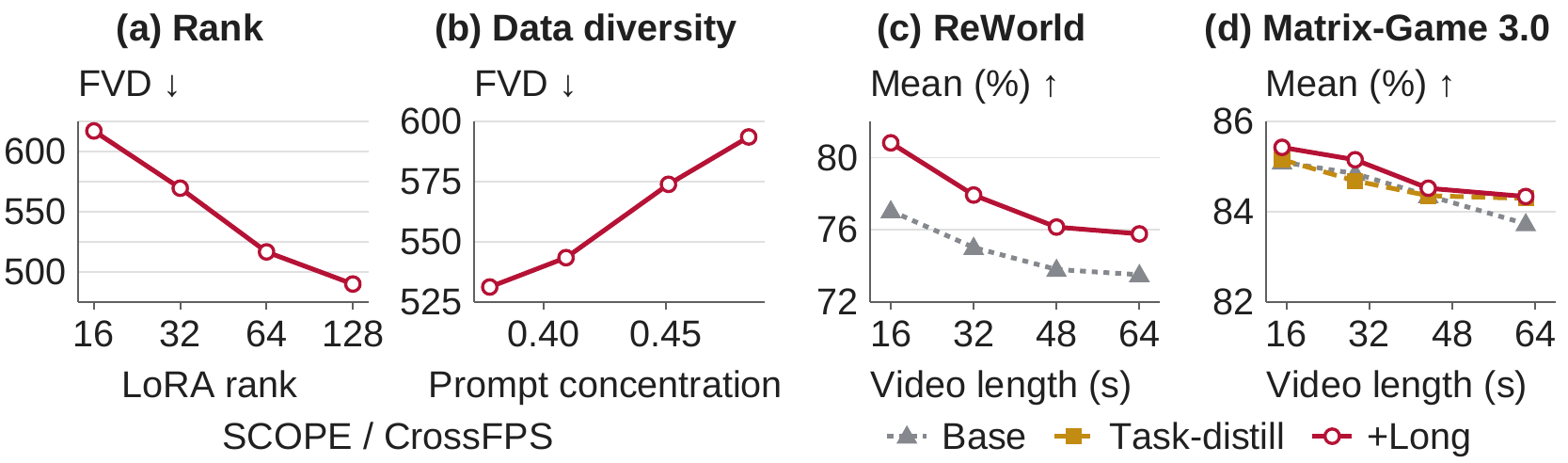}
    \caption{\textbf{Rank, data diversity, and long-context transfer.}
    (a--b) Higher rank and more diverse prompts (lower concentration) improve
    FVD after transfer.
    (c--d) Mean of seven VBench dimensions after transfer to ReWorld and
    Matrix-Game 3.0, respectively. Long-context comparisons are within each
    model. Our transferred long-context LoRA improves quality during long AR rollouts.}
    \label{fig:ablation-transfer-curves}
    \label{fig:rank-ablation}
    \label{fig:data-ablation}
    \label{fig:long-context-transfer}
    \label{fig:reworld-vbench-length}
    \label{fig:matrix-game-vbench-length}
\end{figure}

\subsection{Long-Context Distillation and Transfer}
\label{sec:experiments-long}

We evaluate the transfer of a long-context LoRA distilled on an AR Wan base
model to two AR world models, ReWorld and Matrix-Game 3.0, without downstream
training.

% Effective scores and author corrections: assets/figures/long-context-vbench.json
% and assets/figures/long-context-author-updates.json.
\begin{table}[!htbp]
    \centering
    \caption{\textbf{Long-context transfer to ReWorld and Matrix-Game 3.0.}
    Seven video-intrinsic VBench~\citep{huang2024vbench} dimensions, expressed
    as percentages at each model's longest tested rollout. Best quality
    scores are bolded within each model group.}
    \label{tab:long-context-transfer}
    \label{tab:reworld-vbench-64s}
    \label{tab:matrix-game-vbench-62s}
    \small
    \renewcommand{\arraystretch}{1.15}
    \setlength{\tabcolsep}{3.2pt}
    \begin{adjustbox}{max width=\textwidth}
    % Generated by assets/figures/src/build_long_context_vbench.py.
\begin{tabular}{lcccccccc}
\toprule
Model & \makecell{Imaging\\quality $\uparrow$} & \makecell{Aesthetic\\quality $\uparrow$} & \makecell{Subject\\consist. $\uparrow$} & \makecell{Background\\consist. $\uparrow$} & \makecell{Temporal\\flicker. $\uparrow$} & \makecell{Dynamic\\degree} & \makecell{Motion\\smooth. $\uparrow$} & \makecell{Total\\score $\uparrow$} \\
\midrule
\multicolumn{9}{l}{\textit{ReWorld}} \\[2pt]
ReWorld-base & 33.83 & 37.18 & 63.48 & \textbf{88.87} & 95.29 & 100.00 & 95.89 & 73.51 \\
ReWorld +Long & \textbf{45.41} & \textbf{44.53} & \textbf{66.01} & 80.81 & \textbf{95.75} & 100.00 & \textbf{97.89} & \textbf{75.77} \\
\midrule
\multicolumn{9}{l}{\textit{Matrix-Game 3.0}} \\[2pt]
Matrix-Game-base & 70.76 & 50.51 & 82.72 & 91.10 & 93.69 & 100.00 & 97.34 & 83.73 \\
Matrix-Game-distill & \textbf{74.59} & 51.91 & \textbf{83.35} & 91.39 & 91.93 & 100.00 & 96.90 & 84.30 \\
Matrix-Game +Long & 70.43 & \textbf{54.25} & 82.16 & \textbf{91.51} & \textbf{94.12} & 100.00 & \textbf{97.90} & \textbf{84.34} \\
\bottomrule
\end{tabular}

    \end{adjustbox}
\end{table}

\paragraph{Transfer to ReWorld.}
On ReWorld~\citep{chen2026reworld}, trained on approximately 8\,s windows,
we compare 24-step native inference with four-step +Long over 16--64\,s,
using matched prompts and camera trajectories. At 64\,s, +Long raises the
seven-dimension mean from $73.51$ to $75.77$, improving visual quality and
temporal scores with lower background consistency
(\cref{tab:long-context-transfer}, upper group). Its mean exceeds the base
model at all tested lengths, up to $8\times$ the training duration
(\cref{fig:reworld-vbench-length}c). See Appendix~\ref{sec:appendix-reworld-qualitative}
for qualitative examples.

\paragraph{Transfer to Matrix-Game 3.0.}
We transfer the same +Long adapter to Matrix-Game 3.0~\citep{wang2026matrixgame3}
and compare 50-step native inference, official three-step task-specific
distillation, and four-step +Long under matched inputs and camera actions.
At 62.18\,s, their seven-dimension means are $83.73$, $84.30$, and $84.34$,
respectively (\cref{tab:long-context-transfer}, lower group).
Across tested lengths, +Long is competitive with
task-specific distillation without Matrix-Game training, with
metric-dependent trade-offs (\cref{fig:matrix-game-vbench-length}d).
Qualitative examples appear in Appendix~\ref{sec:appendix-matrix-game}.

\FloatBarrier

\section{Discussion and Limitations}
\label{sec:discussion}

Once-for-all reuse requires compatible descendants of each base model.
Long-context transfer requires existing causal AR inference, since LoRA
updates alone do not change attention masks. Transfer quality involves
task-dependent trade-offs. CFG LoRA
weights provide approximate guidance control and may require adjustment
after transfer.

\section{Conclusion}
\label{sec:conclusion}

We presented \method, which distills CFG, few-step sampling, and
long-context error correction into reusable LoRAs once per backbone family.
These adapters enable training-free transfer to compatible downstream models
with adjustable guidance. Experiments demonstrate effective acceleration and
improved long-video quality across tasks, reducing the need for per-target
distillation.

\par\endgroup

\subsection*{AI use statement}
We used large language models to improve the clarity and readability of the
manuscript, and AI agents to assist with experimental workflows. The authors
are responsible for verifying all AI-assisted work and take full responsibility
for the methods, results, and final content of this paper.

\subsection*{Ethics statement}
This work focuses on improving the efficiency and reuse of video generation
models. We do not anticipate ethical concerns specific to our distillation
framework beyond those associated with the underlying generative models,
including potential misuse for misleading content and inherited biases.
We encourage responsible use in accordance with the licenses and usage
policies of the underlying models and datasets.

\subsection*{Reproducibility statement}
We will publicly release all code and artifacts developed for this work,
including trained LoRA checkpoints, training and evaluation configurations,
and scripts needed to reproduce our experiments. See
Appendix~\ref{sec:appendix-implementation} for implementation details.

{
  \setlength{\bibsep}{3pt}
  \bibliography{reference}
  \bibliographystyle{plain}
}

\clearpage
\appendix
% NV single-column appendix pagination.
% Shared appendix; preserve natural spacing around full-width figures.
% Author requirement: all unresolved work must be visible in red via \TODO{...}.
\raggedbottom
\makeatletter
\providecommand{\theHALG@line}{}
\renewcommand{\theHALG@line}{\thealgorithm.\arabic{ALG@line}}
\makeatother
\section{Implementation Details}
\label{sec:appendix-implementation}

\subsection{CFG-Only LoRA Training}
\label{sec:appendix-cfg-algorithm}

\paragraph{Guided flow regression.}
The student and teacher use the same backbone and attention mode. Given a
noisy state $z_t$ and prompt $c$, the teacher forms the guided flow using
\cref{eq:cfg-teacher}. The negative-prompt branch supplies
$v_\varnothing$; ``unconditional'' therefore denotes the configured
negative condition. The student predicts the target in one conditional
pass.

\paragraph{Training-state construction.}
A captioned clean video latent is noised at a sampled scheduler timestep,
and teacher predictions are evaluated online. We write
$z_t=\mathcal N_t(x,\epsilon)$, where $\mathcal N_t$ denotes the
scheduler's forward-noising operation. For bidirectional T2V, one timestep
is shared by the whole clip, with timestep indices sampled between
$2\%$ and $98\%$ of the training schedule.

\begin{algorithm}[H]
\caption{CFG-only LoRA distillation}
\label{alg:cfg-lora}
\begin{algorithmic}[1]
\Require Frozen teacher $F_{\theta_0}$, trainable adapter $\phi_{\mathrm{cfg}}$,
teacher scale $w_{\mathrm{train}}$, captioned latents, optimizer
\For{each training iteration}
    \State Sample captioned latent $(x,c)$, timestep $t$, and noise $\epsilon$
    \State $z_t\gets\mathcal N_t(x,\epsilon)$
    \State Evaluate frozen $v_c\gets F_{\theta_0}(z_t,t,c)$ and
    $v_\varnothing\gets F_{\theta_0}(z_t,t,\varnothing)$
    \State $v_{\mathrm{cfg}}\gets v_\varnothing+
    w_{\mathrm{train}}(v_c-v_\varnothing)$
    \State $v_s\gets F_{\theta_0\oplus\phi_{\mathrm{cfg}}}(z_t,t,c)$
    \State Compute the flow-regression loss
    \State Backpropagate only through $v_s$; clip gradients; update $\phi_{\mathrm{cfg}}$
\EndFor
\State \Return $\phi_{\mathrm{cfg}}$; deploy with the native schedule and distilled CFG
\end{algorithmic}
\end{algorithm}

\subsection{Few-Step LoRA Training}
\label{sec:appendix-fewstep-algorithm}

Few-step LoRA training follows the distribution-matching objective of
DMD2~\citep{yin2024dmd2}, with a frozen real-score teacher, a generator
adapter $\phi_{\mathrm{step}}$, and a trainable fake-score adapter $\psi$.

\paragraph{Distribution-matching update.}
Given a student sample $x_\phi$, re-noise its detached value at an
independently sampled score timestep $t$. The timestep is warped as
$u\mapsto su/[1+(s-1)u]$ and clamped to $[0.02,0.98]$ of the training
time range. Let $\widehat x_\psi$ and $\widehat x_{\mathrm{real}}^{(w)}$
be the fake-score and guided real-score clean predictions at this state.
The code normalizes their difference within each temporal block $b$:
\begin{equation}
\begin{aligned}
    n_b&=\operatorname{mean}_{f\in b,C,H,W}
        |x_\phi-\widehat x_{\mathrm{real}}^{(w)}|,\\
    g_b&=\operatorname{nan\_to\_num}
        \!\left[(\widehat x_\psi-\widehat x_{\mathrm{real}}^{(w)})/n_b\right],\\
    \mathcal L_G&=\tfrac12\operatorname{mean}
        \left\|x_\phi-\operatorname{sg}[x_\phi-g]\right\|^2.
\end{aligned}
\label{eq:app-dmd}
\end{equation}
The detached target makes the generator gradient proportional to $g$;
gradients do not pass through either score network in this update.
For the fake-score update, another detached student sample is noised and
the fake-score adapter predicts the flow target $\epsilon-x_\phi$:
\begin{equation}
    \mathcal L_D=\operatorname{mean}
    \left\|v_\psi(\mathcal N_t(\operatorname{sg}[x_\phi],\epsilon),t,c)
                 -(\epsilon-\operatorname{sg}[x_\phi])\right\|^2.
    \label{eq:app-critic}
\end{equation}
The selected flow-loss path has no adversarial discriminator term.

\begin{algorithm}[H]
\caption{Wan2.2 few-step LoRA distillation with a learned fake score}
\label{alg:fewstep-lora}
\begin{algorithmic}[1]
\Require Frozen real-score teacher, generator adapter $\phi_{\mathrm{step}}$,
fake-score adapter $\psi$, four-step schedule, update ratio $R=5$
\For{iteration $j=0,\ldots,J-1$}
    \State Sample a training prompt $c$
    \If{$j\bmod R=0$}
        \State Generate a student sample with the four-step schedule
        \State Re-noise the detached clean prediction at a random score timestep
        \State Evaluate conditional fake score and CFG-guided frozen real score
        \State Form detached $g$ and $\mathcal L_G$ using \cref{eq:app-dmd}
        \State Accumulate gradients for $\phi_{\mathrm{step}}$ only
    \EndIf
    \State Generate a separate student sample without gradients
    \State Re-noise it; compute $\mathcal L_D$ using \cref{eq:app-critic}
    \State Accumulate gradients for $\psi$ only
    \State Clip and apply the scheduled generator update and the fake-score update
\EndFor
\State \Return $\phi_{\mathrm{step}}$; discard the fake-score network for inference
\end{algorithmic}
\end{algorithm}

\subsection{Optimization and Deployment}
\label{sec:appendix-training-settings}

Unless varied in the rank ablation, we train each CFG, few-step, and
long-context LoRA at rank $r=128$. The CFG and few-step branches are trained
separately. \Cref{tab:app-training-configs} summarizes the Wan reference
optimization settings.

\begin{table}[H]
\centering
\caption{\textbf{Wan training rank and reference optimization settings.}
All branches are trained at rank 128.
Batch sizes are per process.
For DMD, the generator updates once per five fake-score updates.
The CFG column reports the reference CFG-only optimization settings.}
\label{tab:app-training-configs}
\small
\setlength{\tabcolsep}{4pt}
\begin{tabularx}{\textwidth}{@{}Xccc@{}}
\toprule
Setting & CFG-only, Wan2.2 & Few-step, Wan2.2 & Few-step, Wan2.1\\
\midrule
LoRA training rank $r$ & 128 & 128 & 128\\
Fake-score LoRA & None & Yes & Yes\\
Generator / critic LR & $10^{-5}$ / --- & $10^{-5}$ / $2\!\times\!10^{-6}$ & $10^{-5}$ / $2\!\times\!10^{-6}$\\
AdamW $(\beta_1,\beta_2)$ & $(0,0.999)$ & $(0,0.999)$ & $(0,0.999)$\\
Weight decay & 0 & 0 & 0.01\\
LoRA dropout & 0 & 0 & 0\\
Per-process batch / accumulation & 1 / 1 & 1 / 1 & 1 / 1\\
Generator EMA & Off & 0.99 from step 200 & Off\\
Standard teacher CFG $w$ & 5 & 4 & 5\\
Latent frames $\times C\times H\times W$ & $32\!\times\!48\!\times\!22\!\times\!40$
 & $32\!\times\!48\!\times\!44\!\times\!80$ & $21\!\times\!16\!\times\!60\!\times\!104$\\
Sampling steps & 50 & 4 & 4\\
\bottomrule
\end{tabularx}
\end{table}

Training uses mixed precision, FSDP, and gradient checkpointing. CFG
regression uses FP32 targets and residuals, with gradient clipping at norm 10.
A DMD checkpoint iteration counts fake-score updates; the generator updates
only at iterations divisible by five.

At deployment, we merge the CFG and few-step updates into the downstream
weights as in \cref{eq:cfg-step-composition}, preserving target-specific modules.
We fix $\lambda_{\mathrm{step}}=1$ and adjust $\lambda_{\mathrm{cfg}}$
for each downstream task.

\clearpage
\section{Additional Acceleration Comparisons}
\label{sec:appendix-qualitative}

\begin{figure}[H]
    \centering
    \includegraphics[width=\textwidth]{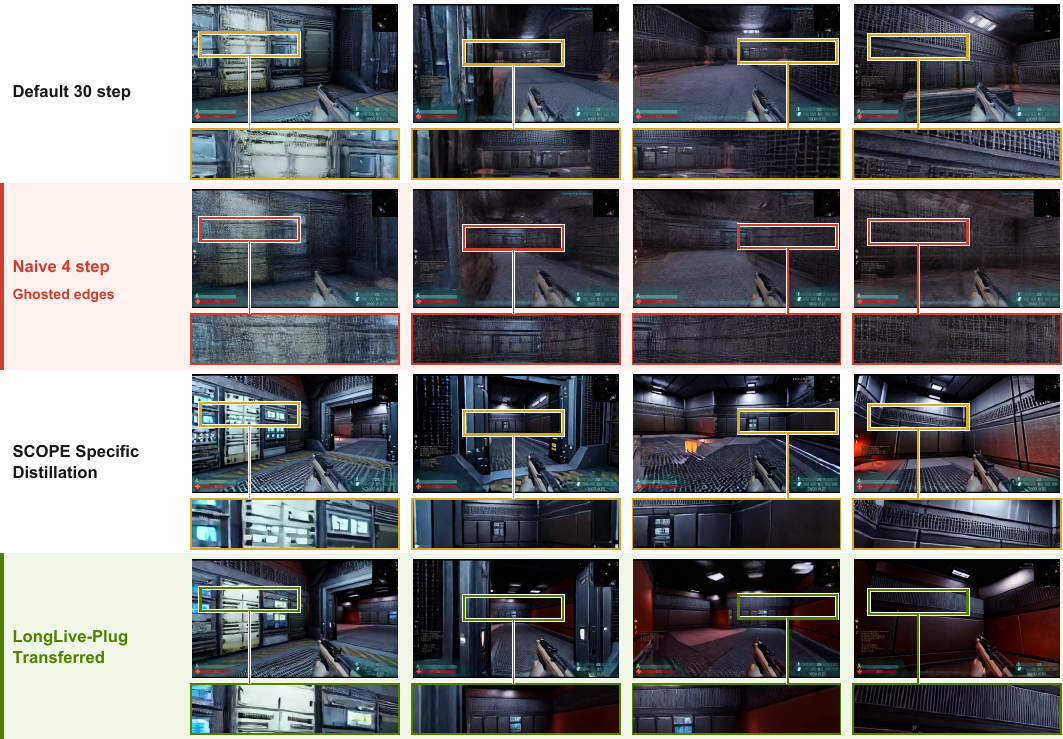}
    \caption{\textbf{Keyframe comparison on SCOPE.}
    Four matched keyframes from an 81-frame sequence at 20 fps.
    Rows compare default 30-step inference, naive four-step sampling,
    SCOPE-specific distillation, and transferred LoRAs, using the same case
    and seed. Boxes mark identical image coordinates across methods;
    the strips below each frame magnify these regions. Red highlights the
    naive four-step row's ghosted wall and door edges, while the distilled
    variants retain clearer boundaries and surface detail.}
    \label{fig:scope-transfer-extended}
\end{figure}

\begin{figure}[H]
    \centering
    \includegraphics[width=\textwidth]{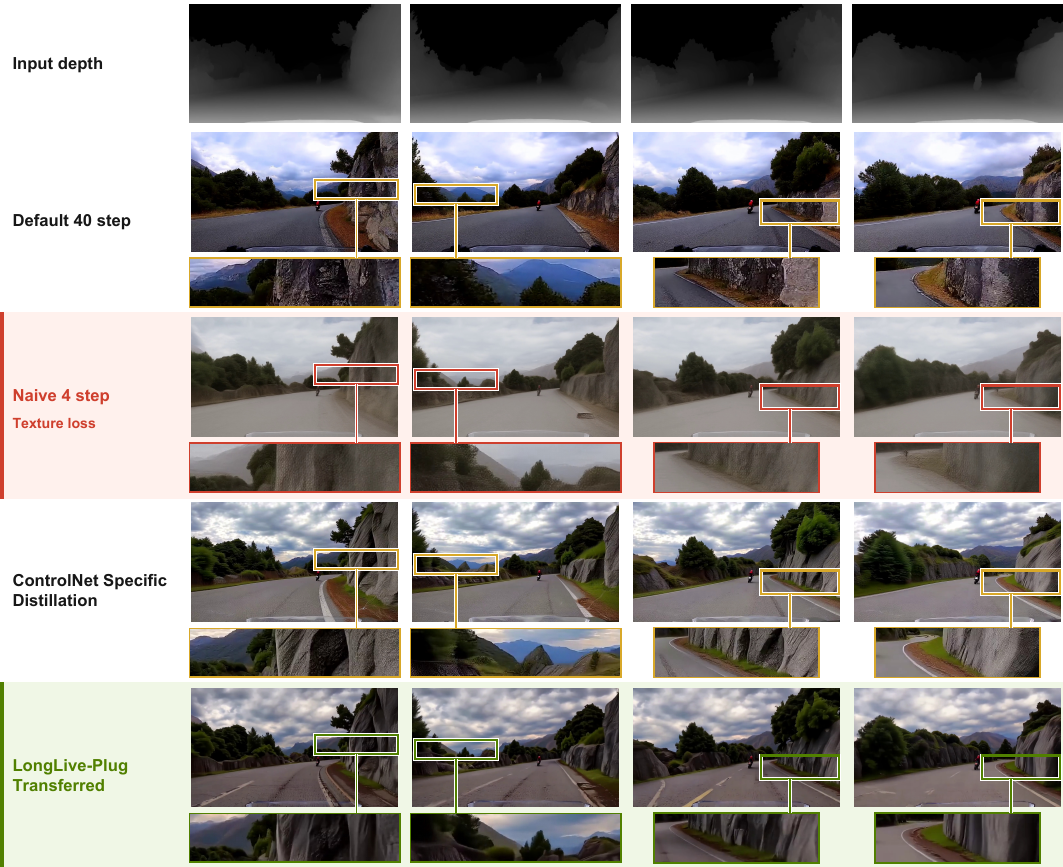}
    \caption{\textbf{Keyframe comparison on Wan2.2-Fun-5B-Control.}
    Four matched keyframes compare input depth and four generation methods.
    Boxes mark identical image coordinates across methods, magnified in the
    strips below each output. Red highlights the naive four-step row's
    smeared rock, foliage, and road details and low contrast. Columns match
    conditioning frame indices (depth at 30 fps, outputs at 24 fps).
    The transferred output uses an additional base-distilled adapter variant.}
    \label{fig:control-transfer-extended}
\end{figure}

This section extends \cref{sec:experiments-main}. We show more cases in
\cref{fig:scope-transfer-extended,fig:control-transfer-extended}.

\clearpage

\clearpage
\section{Complete Transfer Coverage and Additional Cases}
\label{sec:appendix-coverage}

\subsection{Coverage Inventory}
\label{sec:appendix-coverage-audit}

The inventory supporting \cref{sec:experiments-coverage} contains
54 distinct downstream model entries: 24 descendants of Wan2.1-14B,
24 of Wan2.2-TI2V-5B, and six of MiniMax-H3.

Each family uses an adapter distilled on its own base model.
MiniMax-H3 natively supports inference without CFG, so its transfer
experiments omit the CFG LoRA by default. We also train a CFG-only LoRA
on the MiniMax-H3 base model and find that it supports distilled CFG
inference with adjustable guidance (\cref{fig:app-h3-cfg-style}).

\begin{table}[!htbp]
    \centering
    \caption{\textbf{Recorded transfer coverage across backbone families and tasks.}
    Every listed Wan-model entry is evaluated for four-step and CFG transfer.
    The two panels share the same backbone axis.}
    \label{tab:transfer-coverage}
    \fontsize{7.3}{9.2}\selectfont
    \setlength{\tabcolsep}{4pt}
    \renewcommand{\arraystretch}{1.04}
    \begin{tabularx}{\textwidth}{@{}p{0.135\textwidth}*{4}{>{\raggedright\arraybackslash}X}@{}}
\toprule
\textbf{Backbone} & \textbf{World models} & \textbf{Robotics} & \textbf{ControlNet / structure} & \textbf{Camera / trajectory} \\
\midrule
\textbf{\makecell[l]{Wan2.1-14B}} & FantasyWorld \citep{dai2025fantasyworld}; LongVie 2 \citep{gao2025longvie2}; Micro-World I2W \citep{amd2025microworldi2w} & DreamZero \citep{ye2026dreamzero}; ABot-PhysWorld \citep{chen2026abotphysworld}; VERA DROID Planner \citep{li2026veradroidwanplanner14b} & Fun Control \citep{alibabapai2025wan21fun14bcontrol}; TheDenk Dilated ControlNet \citep{karachev2025dilatedcontrolnet} & Fun-V1.1 Control-Camera \citep{alibabapai2025wan21funv1114bcontrolcamera}; Wan-Move \citep{chu2025wanmove14b480p}; ATI \citep{wang2025atianytrajectoryinstruction}; NeoVerse \citep{yang2026neoverse}; Vista4D \citep{lin2026vista4d} \\[5pt]
\textbf{\makecell[l]{Wan2.2-\\TI2V-5B}} & Matrix-Game 3.0 \citep{wang2026matrixgame3}; DreamX-World-5B AR \citep{dreamxteam2026dreamxworld5bar}; SCOPE \citep{tong2026scope}; ReactiveGWM \citep{wang2026reactivegwm}; Spatia \citep{zhao2025spatia} & Boundless World Model \citep{bwmteam2026boundlessworldmodel}; Fast-WAM LIBERO \citep{yuan2026fastwamlibero}; Motus Stage-1 VGM \citep{bi2025motusstage1vgm} & Fun Control \citep{alibabapai2025wan22fun5bcontrol} & Fun Control-Camera \citep{alibabapai2025wan22fun5bcontrolcamera}; FlashMotion \citep{li2026flashmotion}; FrameINO v1.6 \citep{wang2025frameino5bv16} \\[5pt]
\textbf{\makecell[l]{MiniMax-H3}} & H3-World \citep{chen2026h3world}; Code World Model \citep{chen2026codeworldmodel}; SolarWM-H3 \citep{huang2026solarwmh3} & --- & Fun ControlNet-Union \citep{alibabapai2026minimaxh3funcontrolnetunion} & SolarWM-H3 \citep{huang2026solarwmh3} \\[5pt]
\bottomrule
\end{tabularx}
\par\vspace{7pt}
\begin{tabularx}{\textwidth}{@{}p{0.135\textwidth}*{4}{>{\raggedright\arraybackslash}X}@{}}
\toprule
\textbf{Backbone} & \textbf{Editing / restoration} & \textbf{Subject / avatar} & \textbf{Audio / RGBA outputs} & \textbf{Domain / style / quality} \\
\midrule
\textbf{\makecell[l]{Wan2.1-14B}} & VideoCoF \citep{yang2025videocof14b}; SAMA-14B \citep{zhang2026sama14b} & Stand-In \citep{xue2025standint2v14b}; MAGREF \citep{deng2025magref}; MagicTryOn \citep{li2025magictryon14bv1}; InfiniteTalk \citep{yang2025infinitetalk}; MultiTalk \citep{kong2025multitalk}; FantasyTalking \citep{wang2025fantasytalking}; AniCrafter \citep{niu2025anicrafter} & Wan-Alpha v1/v2 \citep{dong2025wanalphav1v2} & Index-AniSora V2.0 \citep{jiang2024indexanisorav20,indexteam2025anisora2} \\[5pt]
\textbf{\makecell[l]{Wan2.2-\\TI2V-5B}} & Fun InP \citep{alibabapai2025wan22fun5binp}; Kiwi-Edit \citep{lin2026kiwiedit}; StreamingEffect \citep{song2026streamingeffect} & TalkVerse-5B \citep{wang2025talkverse5b} & Ovi \citep{low2025ovi}; OmniCustom \citep{li2026omnicustom}; NAVA \citep{ji2026nava}; UniAVGen \citep{zhang2025uniavgen} & Loomis Painter LoRA \citep{pobitzer2025loomispainterlora}; Aether action/VFX LoRAs \citep{sallstrom2025aetherpunch}; VideoGPA DPO LoRA \citep{du2026videogpadpolora}; PhysRVG \citep{zhang2026physrvg} \\[5pt]
\textbf{\makecell[l]{MiniMax-H3}} & LineartAnime \citep{diffsynth2026lineartanime} & Viggle-Animate \citep{viggle2026animate} & --- & --- \\[5pt]
\bottomrule
\end{tabularx}

\end{table}

\clearpage
\subsection{Additional Wan Transfer Examples}

\begin{figure}[H]
    \centering
    \includegraphics[width=\textwidth]{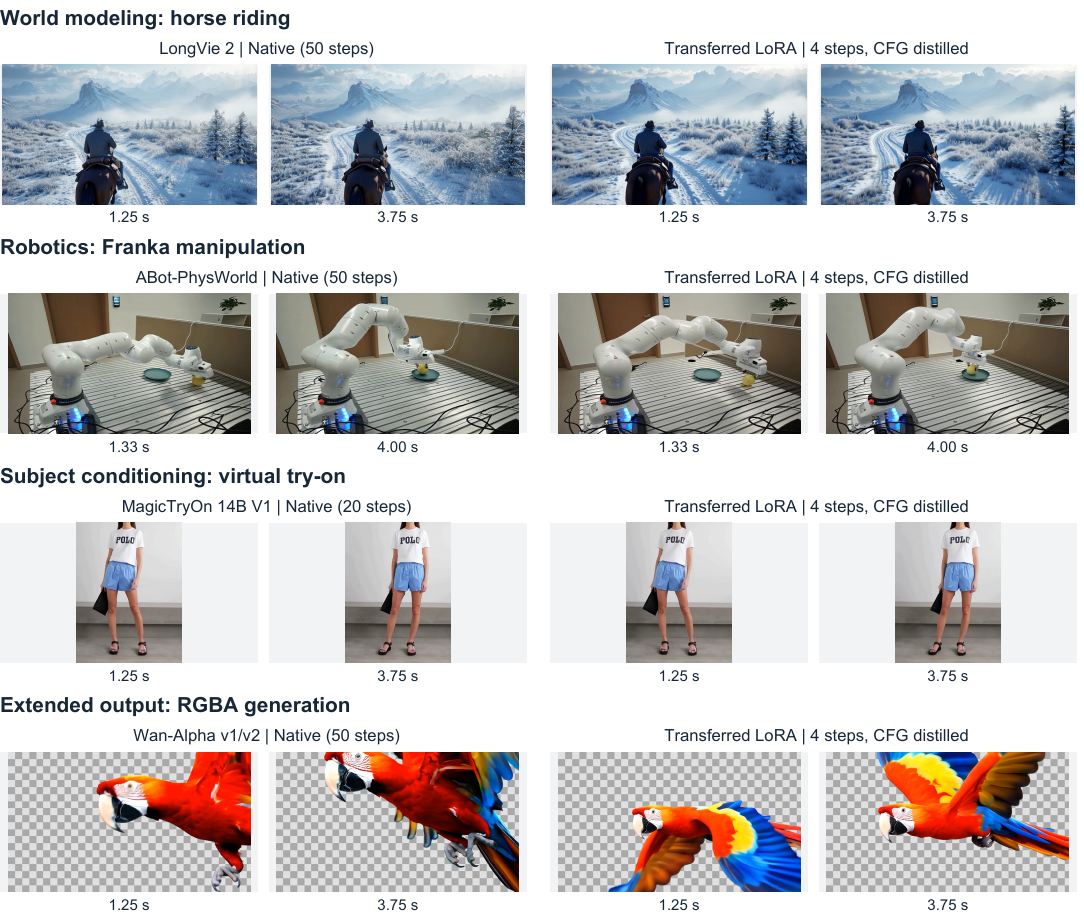}
    \caption{\textbf{Additional Wan2.1-14B transfer cases.}
    LongVie 2, ABot-PhysWorld, MagicTryOn, and Wan-Alpha illustrate world
    modeling, robotics, subject conditioning, and RGBA output.
    Each row compares two native frames (left) with the same two timestamps
    after four-step transfer (right). Inputs and task conditions are paired
    in the source report. The checkerboard is part of the Wan-Alpha preview,
    not a measurement of alpha-channel accuracy. Changes in appearance
    and motion remain visible after transfer.}
    \label{fig:app-wan21-transfer}
\end{figure}

\clearpage
\begin{figure}[H]
    \centering
    \includegraphics[width=\textwidth]{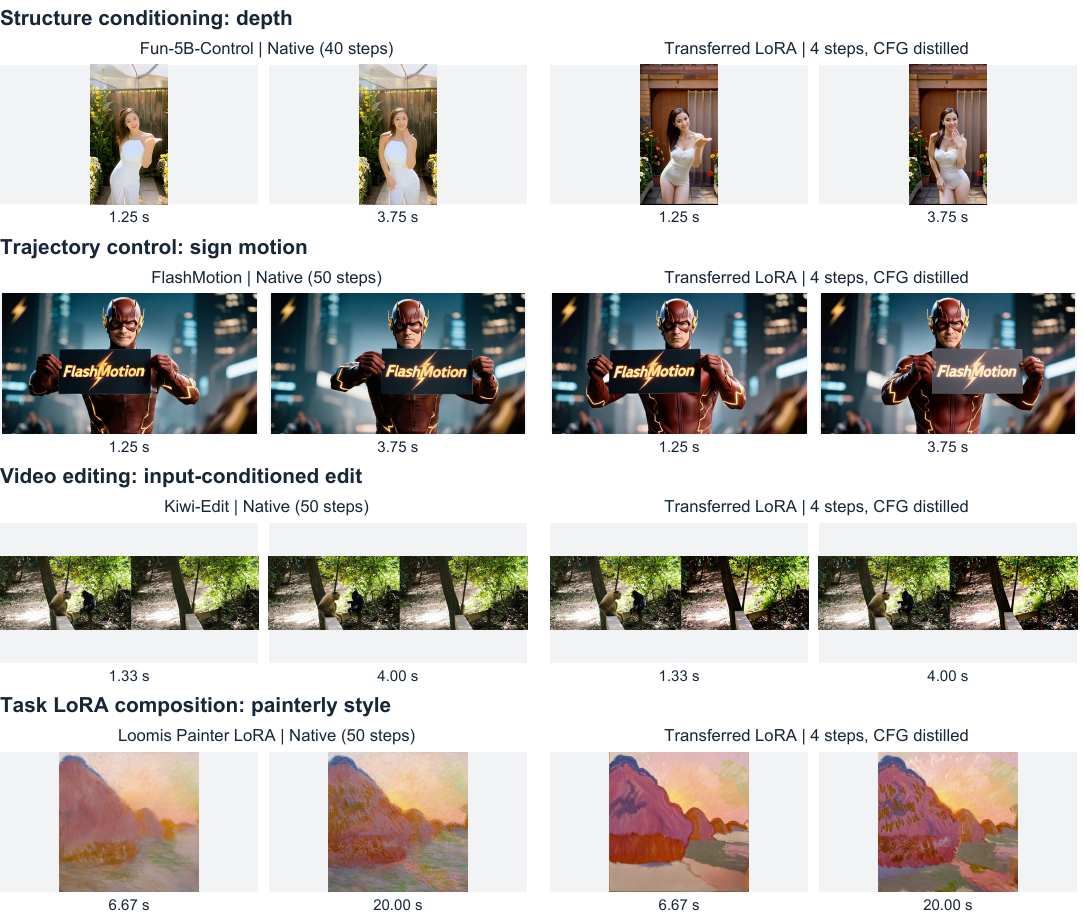}
    \caption{\textbf{Additional Wan2.2-TI2V-5B transfer cases.}
    Depth-conditioned Fun Control, FlashMotion, Kiwi-Edit, and Loomis Painter
    cover structure, trajectory, editing, and style adaptation. The native
    and transferred columns show identical frame indices for each paired
    task input. Native inference uses 50 steps for Kiwi-Edit and Loomis Painter.
    All transferred outputs use four steps
    with CFG distilled into the adapter; downstream conditioning and task
    adapters are retained.}
    \label{fig:app-wan22-transfer}
\end{figure}

\clearpage
\subsection{MiniMax-H3 Transfer and Its Limits}

The H3 comparisons show undistilled multi-step inference on the left (D),
naive four-step Euler in the middle (E4), and four-step inference with
\method on the right (S4). S4 uses fresh re-noising. The downstream
conditions are fixed.
E4 and S4 share the initial noise; D retains the native random-number
generation path, so bitwise-identical initial noise across all three
arms is not established. E4 to S4 changes both the adapter and sampler.
The following examples assess the complete deployment recipe,
not the isolated causal effect of adding a LoRA.

\begin{figure}[H]
    \centering
    \includegraphics[width=\textwidth]{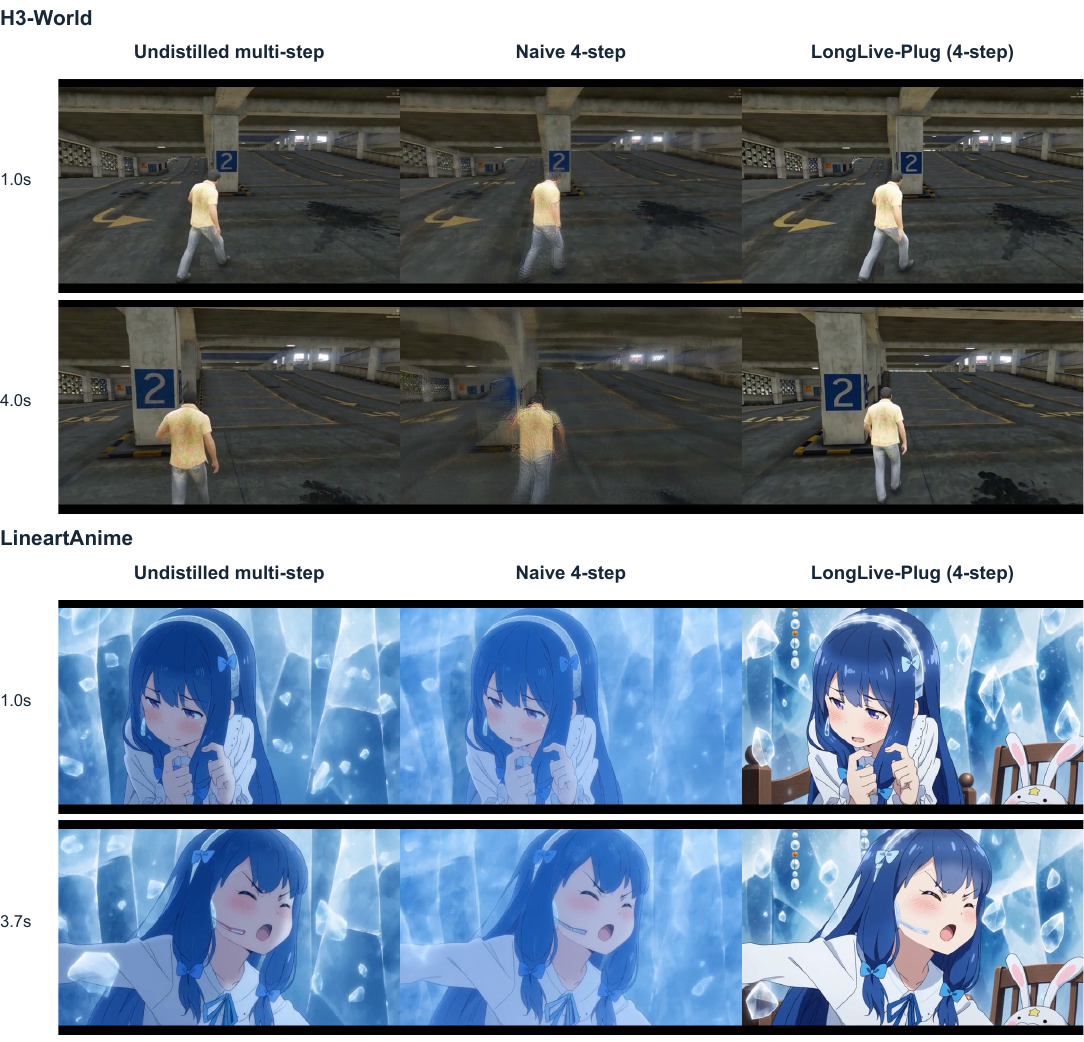}
    \caption{\textbf{H3 task transfer: action control and line-art coloring.}
    \textbf{Left:} undistilled multi-step inference.
    \textbf{Middle:} naive four-step sampling.
    \textbf{Right:} four-step inference with \method. H3-World shows matched frames at 1 and 4 s
    under a forward-action condition. LineartAnime shows frames at 1 s and
    the final available frame (3.71 s). S4 preserves a clearer character
    outline than E4 in these examples, while appearance can differ from D.
    These static frames do not evaluate audio quality or synchronization.}
    \label{fig:app-h3-transfer}
\end{figure}

\clearpage
\begin{figure}[H]
    \centering
    \includegraphics[width=\textwidth]{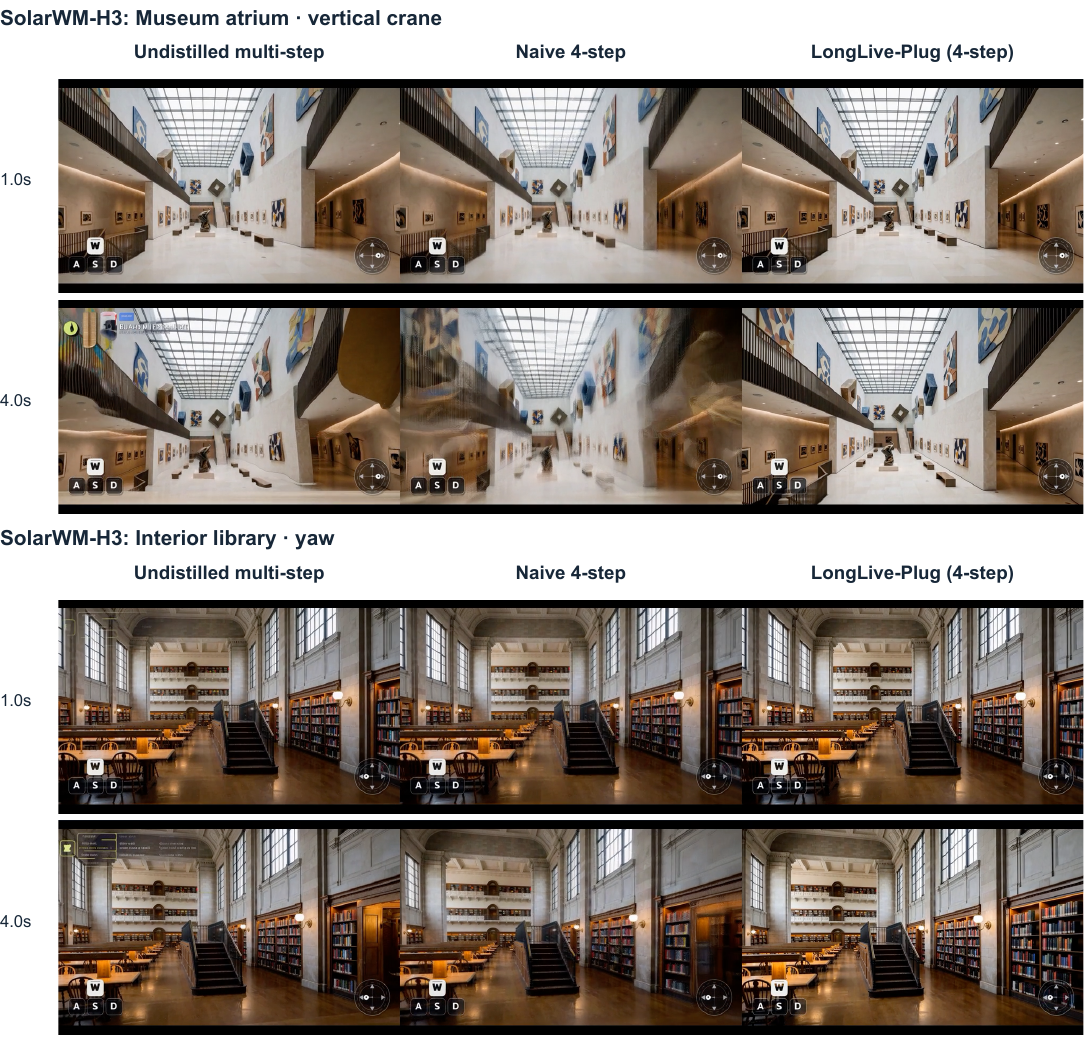}
    \caption{\textbf{H3 camera transfer includes a quality--control trade-off.}
    \textbf{Left:} undistilled multi-step inference.
    \textbf{Middle:} naive four-step sampling.
    \textbf{Right:} four-step inference with \method.
    In the museum-crane example, S4 retains
    clearer architectural detail and a visible upward-camera response.
    In the library-yaw example, S4 remains sharp but its change in framing
    is attenuated relative to D/E4. HUD elements are inherited from the
    supplied anchor images. These cases have no generated-video pose
    regression metric and do not establish precise trajectory adherence.}
    \label{fig:app-h3-camera}
\end{figure}

The ten H3 comparisons are single-seed cases, with no repeated-run
confidence intervals. The native default is a reference operating point,
not ground truth. Similarity to it cannot by itself measure action,
camera, or structure-control accuracy. Runtime accounting also differs
across H3 backends, so timings should only be compared within a task.

\clearpage
\section{Additional CFG Control Experiments}
\label{sec:appendix-cfg}

This section extends \cref{sec:experiments-cfg} with native-schedule CFG-only
control and four-step composition. Displayed LoRA weights scale adapter
residuals and are not calibrated runtime CFG scales.

\subsection{CFG-Only Control with the Native Schedule}
\label{sec:appendix-cfg-alone}

% Keep the two single-case comparisons together at full width.
\begingroup
\setlength{\intextsep}{6pt plus 1pt minus 1pt}
\captionsetup{skip=4pt}

\paragraph{Wan2.2-TI2V-5B.}
The CFG-only experiment uses the native 50-step FlowUniPC schedule,
no few-step adapter, teacher scale $w_{\mathrm{train}}=5$, and runtime CFG $1$.
\Cref{fig:cfg-guidance-base-flower} adds a flower-opening prompt using
the main-text milk-splatter comparison protocol.

\begin{figure}[H]
    \centering
    \includegraphics[width=\textwidth]{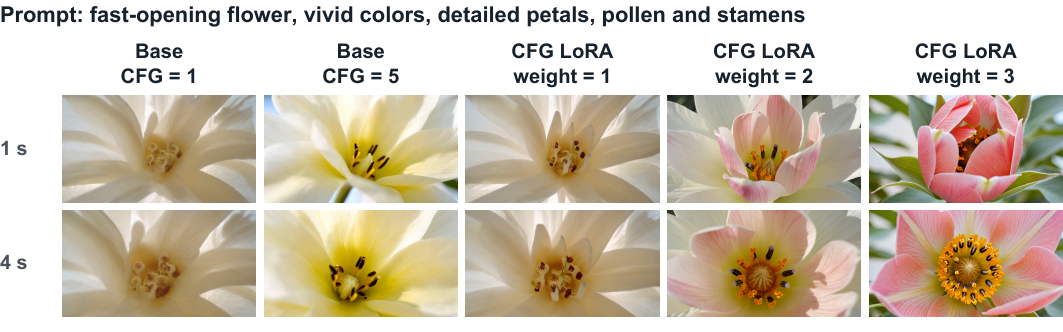}
    \caption{\textbf{Additional Wan CFG-only control example.}
    Native CFG references and CFG LoRA weights $1$, $2$, and $3$ use
    50 sampling steps; LoRA outputs use runtime CFG $1$ with one conditional
    evaluation per step. Matched frames at 1 and 4 s show more pronounced
    flower opening as the adapter weight increases.}
    \label{fig:cfg-guidance-base-flower}
\end{figure}

\paragraph{MiniMax-H3.}
Native CFG and CFG LoRA outputs are compared in
\cref{fig:app-h3-cfg-rooftop,fig:app-h3-cfg-style}.

\begin{figure}[H]
    \centering
    \includegraphics[width=\textwidth]{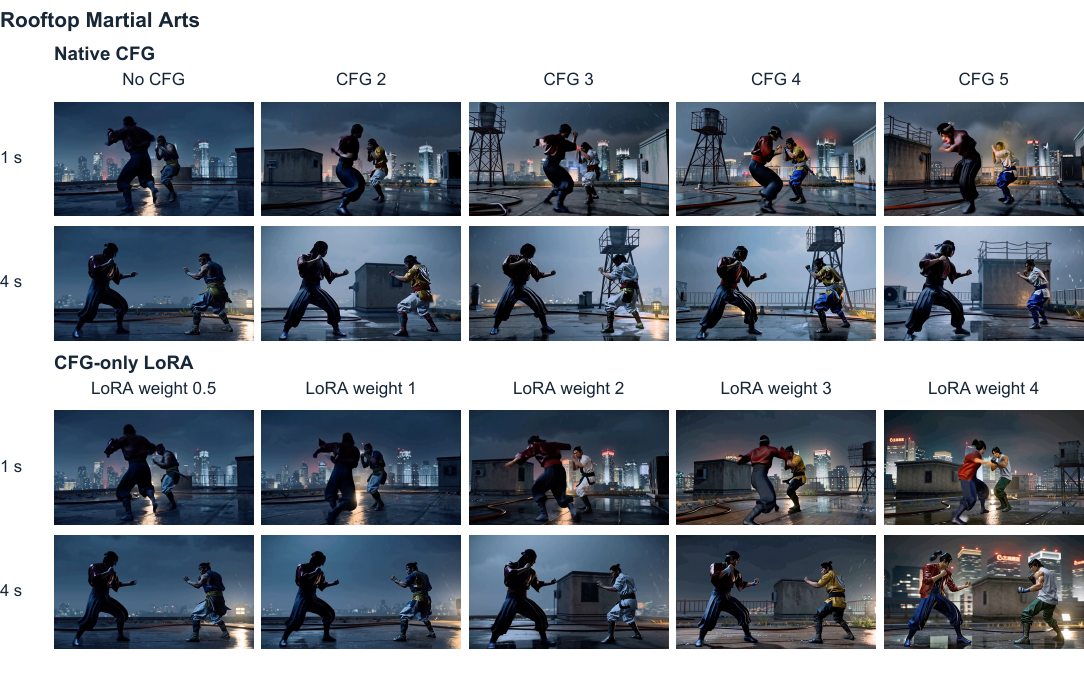}
    \caption{\textbf{Native CFG and CFG-only LoRA on Rooftop Martial Arts.}
    Matched frames at 1 and 4 s compare native CFG (top: no CFG and scales
    $2$--$5$) with CFG LoRA (bottom: weights $0.5$, $1$, $2$, $3$, $4$).
    The prompt, seed, and native schedule are fixed; LoRA outputs use one
    conditional forward pass per step. LoRA weights are not calibrated
    native CFG scales.}
    \label{fig:app-h3-cfg-rooftop}
\end{figure}

\noindent\begin{minipage}{\textwidth}
The MiniMax-H3 CFG-only report covers 34 prompts, seven native reference
conditions, and six adapter weights per prompt (442 videos). The adapter
is trained at CFG scale $3$ and uses no external CFG. The native 50-step
scheduler performs 49 denoising updates (one conditional forward per LoRA
update), generating 124 frames at 24 fps and $1344\times768$ resolution;
video/audio flow shifts are $12/3$.
\Cref{fig:app-h3-cfg-rooftop,fig:app-h3-cfg-style} show Rooftop Martial Arts,
Night village---neutral, and Night village---subtle Van Gogh influence.
Each case places native CFG above CFG LoRA with matched frames at 1 and
4 s and a fixed prompt and seed. Native columns use no CFG and scales
$2$--$5$; adapter weights are $0.5$, $1$, $2$, $3$, and $4$.
\end{minipage}

\begin{figure}[H]
    \centering
    \includegraphics[width=\textwidth]{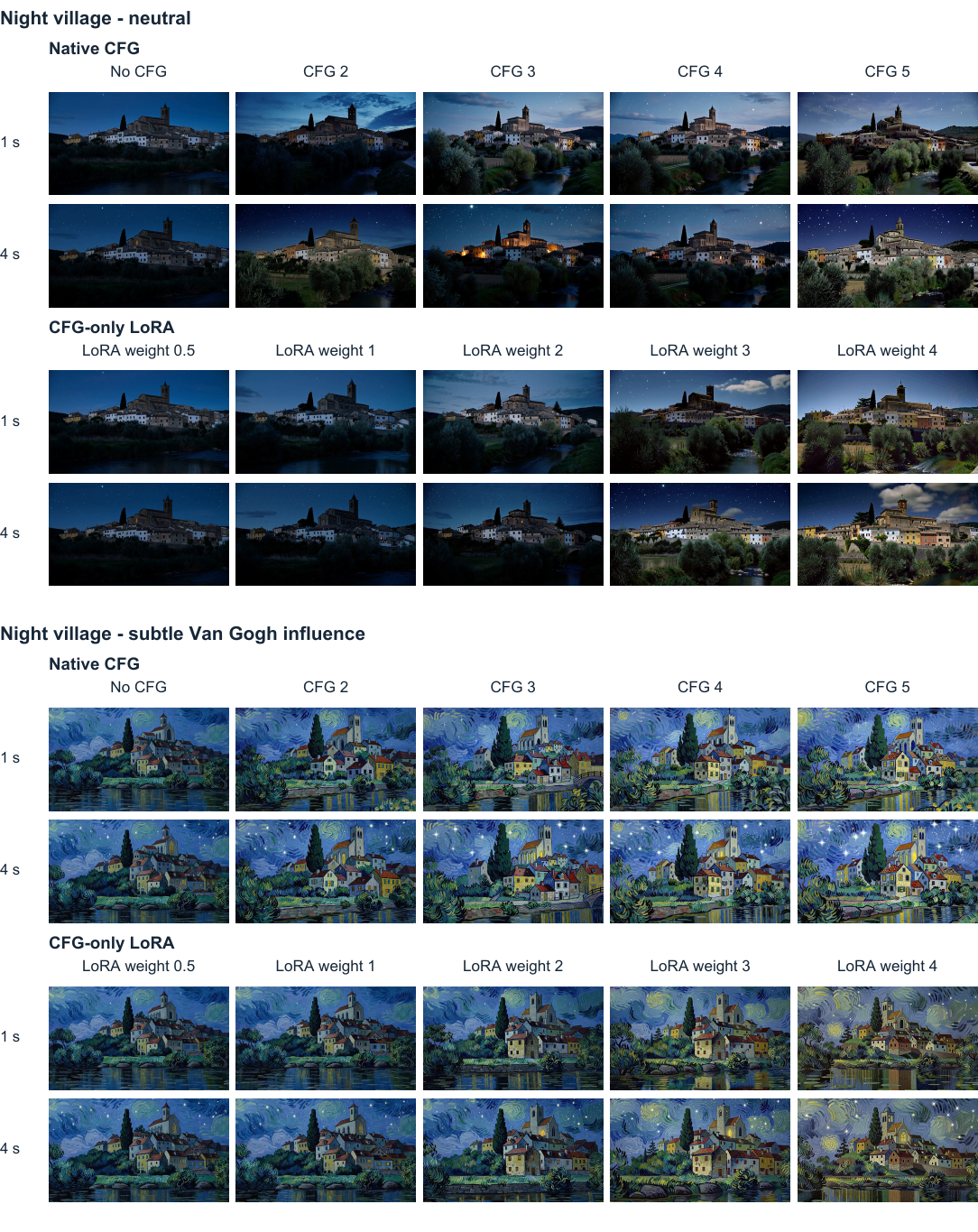}
    \caption{\textbf{Native CFG and CFG-only LoRA on the two night-village cases.}
    Each case places native CFG above CFG LoRA at matched 1 and 4 s frames,
    using the sweeps in \cref{fig:app-h3-cfg-rooftop}. The prompt, seed,
    and native sampling schedule are fixed within each case.}
    \label{fig:app-h3-cfg-style}
\end{figure}

\endgroup

\clearpage
\subsection{Independent CFG Control with a Few-Step Adapter}
\label{sec:appendix-cfg-composition}

We compose the two branches as $\Delta W=\Delta W_{\mathrm{step}}+
\lambda_{\mathrm{cfg}}\Delta W_{\mathrm{cfg}}$. The few-step weight remains
$1$ while $\lambda_{\mathrm{cfg}}\in\{0.5,1,2,3,5\}$ varies. Every output
uses four denoising steps, distilled CFG, and scheduler shift $5$.
Within each sweep, the prompt, seed, conditioning, and action sequence
are fixed. These examples show that the CFG branch remains an effective
semantic control when combined with the few-step branch.

\begin{figure}[H]
    \centering
    \includegraphics[width=\textwidth]{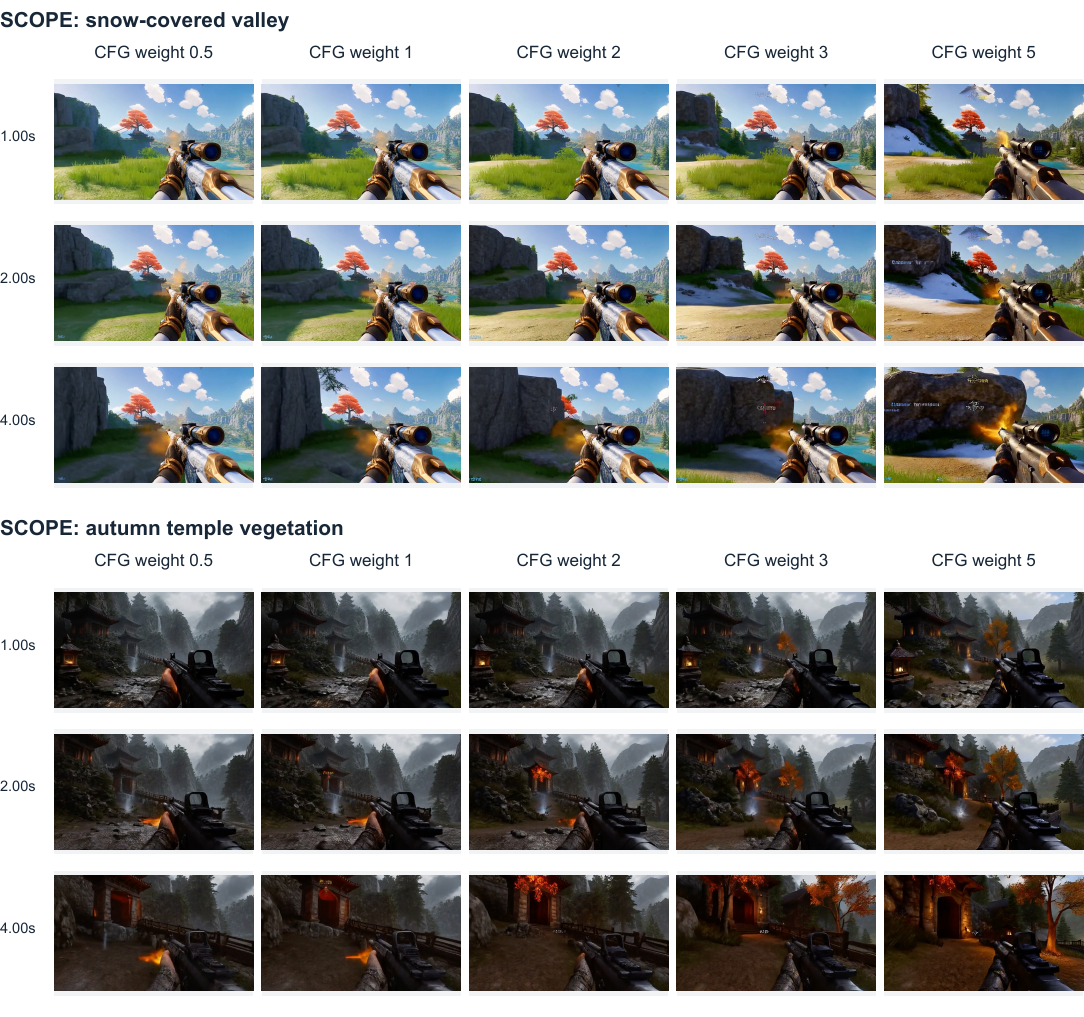}
    \caption{\textbf{CFG-branch control with four-step SCOPE generation.}
    The prompts request snow cover in a mountain valley and autumn
    vegetation around an ancient temple. Both use 81 frames at 20 fps;
    rows show 1, 2, and 4 s. Increasing the CFG-branch weight
    strengthens the snow cover and orange-red foliage, while the few-step
    weight stays at $1$. An excessively large weight ($5$) degrades quality,
    changing geometry and introducing spurious text.}
    \label{fig:app-cfg-composition}
\end{figure}

\clearpage
\begin{figure}[H]
    \centering
    \includegraphics[width=\textwidth]{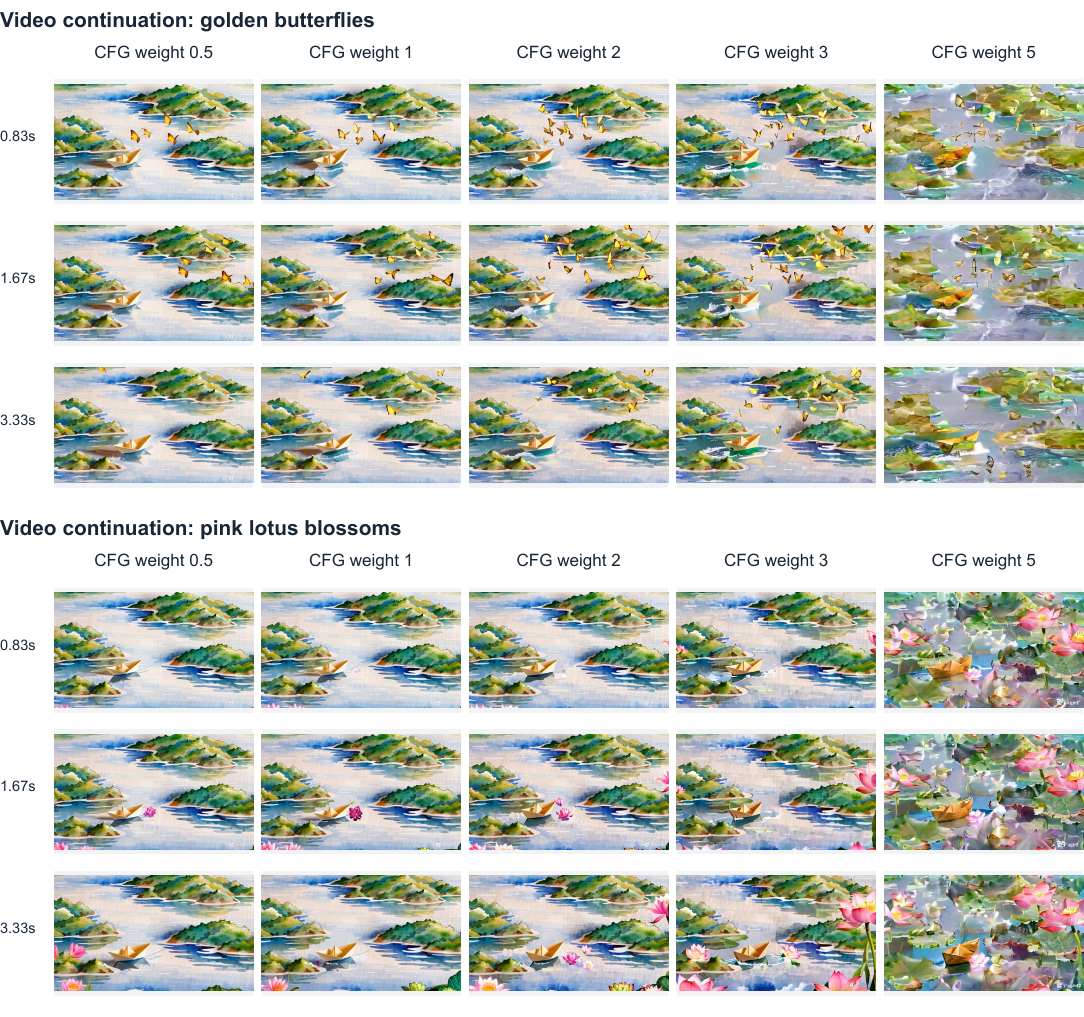}
    \caption{\textbf{CFG-branch control with four-step video continuation.}
    We continue the same watercolor paper-boat input with golden
    butterflies or pink lotus blossoms. The last
    24 input frames condition 81 new frames at 24 fps; displayed times are
    relative to the generated continuation. Weights $2$--$3$ produce more
    visible and persistent butterflies or more prominent lotus blossoms.
    At weight $5$, dense generated content comes with fragmented scenery
    and stronger changes to the boat and islands. The few-step branch
    remains fixed at weight $1$ in every column.}
    \label{fig:app-cfg-continuation}
\end{figure}

\Cref{fig:app-cfg-composition,fig:app-cfg-continuation} show that increasing
the CFG LoRA weight strengthens text guidance, making the requested
attributes more pronounced. However, overly large weights, such as $5$,
degrade visual quality. The CFG LoRA weight should therefore be adjusted
within a reasonable range for each task, balancing text guidance and
visual quality.

\clearpage
\section{Long-Context Qualitative Comparisons}
\label{sec:appendix-reworld-qualitative}

This section accompanies \cref{sec:experiments-long}, with long rollouts
on ReWorld and Matrix-Game 3.0.

\subsection{ReWorld}

\begin{figure}[H]
    \centering
    \includegraphics[width=\textwidth,height=0.68\textheight,keepaspectratio]{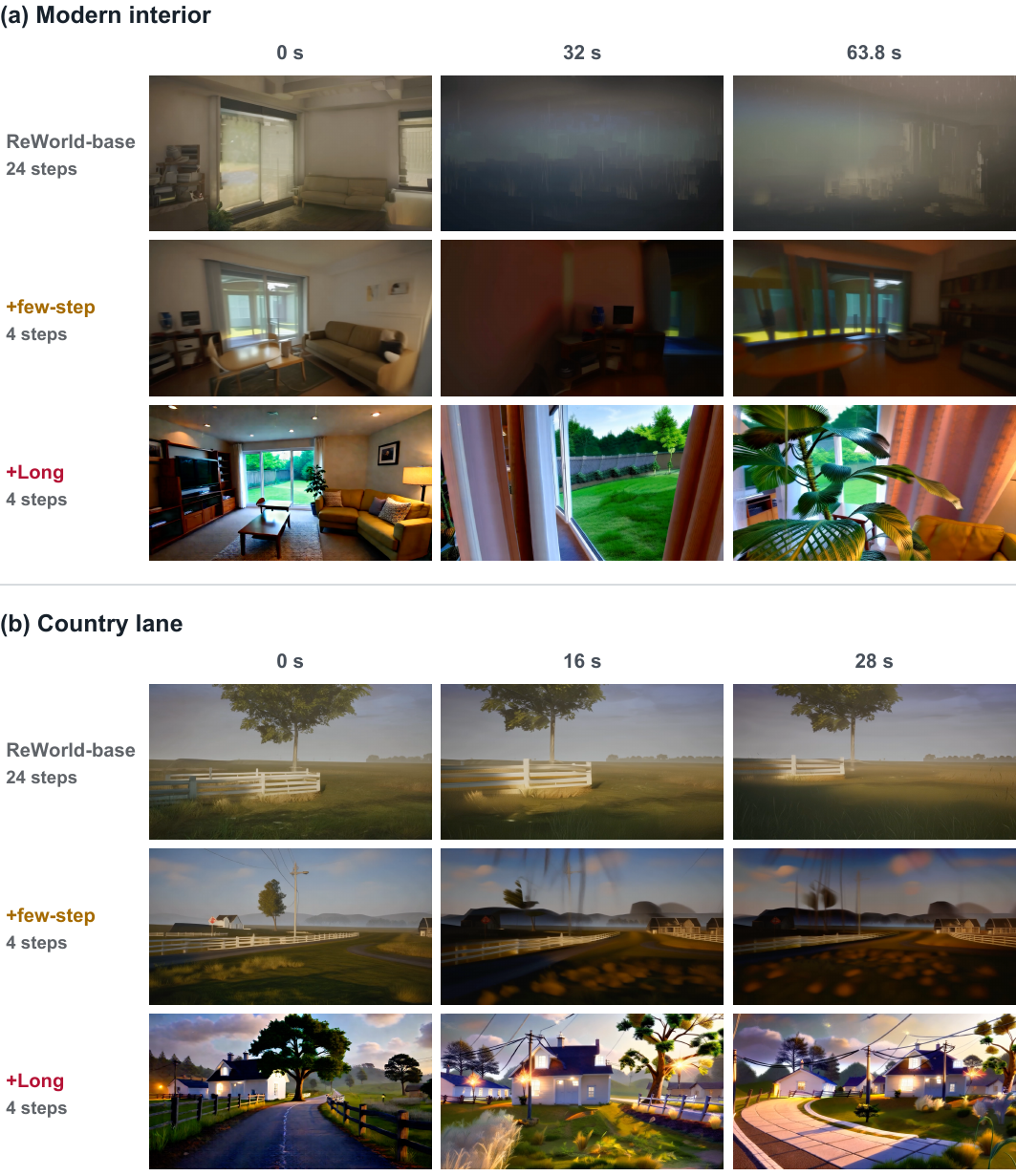}
    \caption{\textbf{Long-rollout qualitative comparisons on ReWorld.}
    Matched frames from a modern interior and a country lane.
    The +Long outputs retain more visible texture and object detail at late times.}
    \label{fig:reworld-long-qualitative}
\end{figure}

\Cref{fig:reworld-long-qualitative} retains two 64\,s cases with matched
prompts, camera trajectories, and initial noise.
Generated layouts can differ across methods.

\clearpage
\subsection{Matrix-Game 3.0}
\label{sec:appendix-matrix-game}

We show two cases. Each rollout has 1,057 frames at 17 fps and
lasts 62.18\,s. All four methods share the
same input image, prompt, seed, and frozen action sequence.

% Stack the two full-width Matrix-Game comparisons on one page.
\begingroup
\setlength{\intextsep}{4pt plus 1pt minus 1pt}
\captionsetup{skip=4pt}
\begin{figure}[H]
    \centering
    \includegraphics[width=\textwidth]{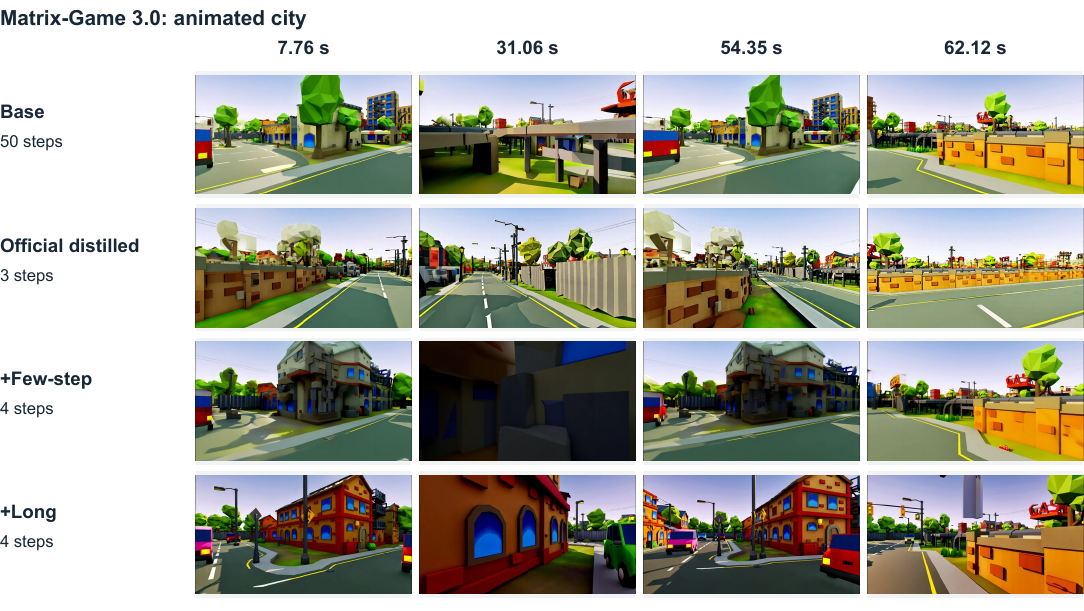}
    \caption{\textbf{Long-context transfer to Matrix-Game 3.0: animated city.}
    Columns show native frames 132, 528, 924, and 1,056 (zero-based),
    spanning early, middle, and late stages of the same 62.18\,s rollout.
    The +Long output retains distinct facade edges and street objects
    at late times.}
    \label{fig:app-matrix-game-city}
\end{figure}

\begin{figure}[H]
    \centering
    \includegraphics[width=\textwidth]{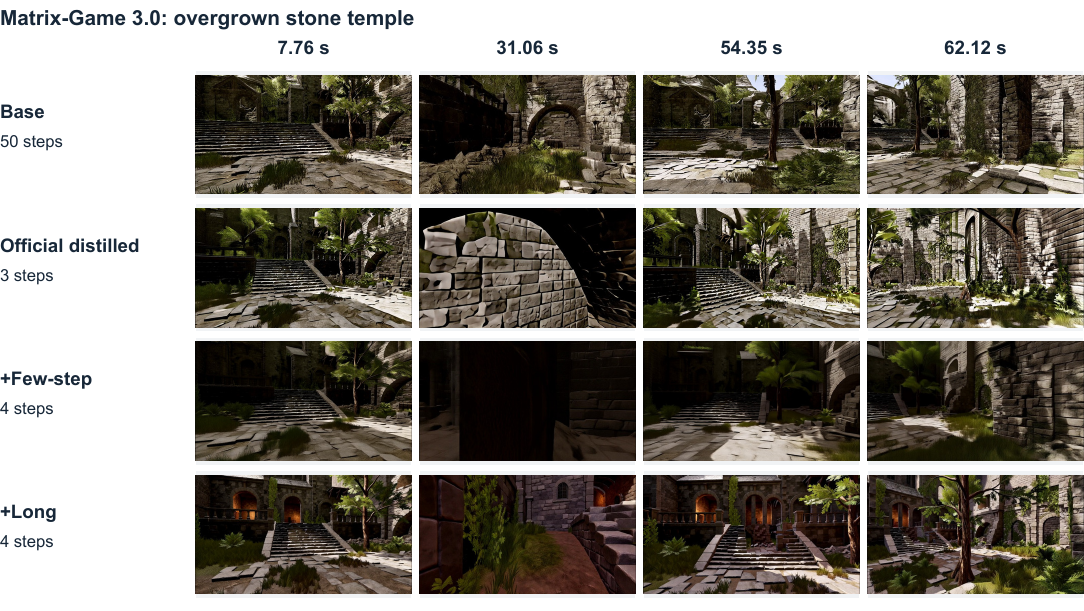}
    \caption{\textbf{Long-context transfer to Matrix-Game 3.0: overgrown temple.}
    The same methods and timestamps as \cref{fig:app-matrix-game-city}
    compare stone architecture and vegetation. The +Long frames preserve
    visible stone-block boundaries, steps, and foliage late in the rollout,
    while appearance and layout vary across methods.}
    \label{fig:app-matrix-game-temple}
\end{figure}
\endgroup
\FloatBarrier

\end{document}